%% file: camera_ready.tex
\documentclass[letterpaper]{article} 
\usepackage[]{aaai24}  

\usepackage{times}  
\usepackage{helvet}  
\usepackage{courier}  
\usepackage[hyphens]{url}  
\usepackage{graphicx} 
\usepackage{natbib}  
\usepackage{caption} 
\input{math_commands.tex}

\usepackage{amsmath}
\usepackage{amssymb}
\usepackage{mathtools}
\usepackage{amsthm,booktabs}
\usepackage{algorithm,algorithmic}

\usepackage{graphicx}
\graphicspath{ {./figures/} }
\usepackage{caption}
\usepackage{subcaption}
\usepackage{extarrows}
\usepackage{multirow,amssymb}
\usepackage{amsmath}
\usepackage{makecell}
\usepackage{enumitem}
\usepackage{pifont}
\usepackage{bm}
\input{pream_bm}

\theoremstyle{plain}
\usepackage{chngcntr}
\newtheorem{theorem}{Theorem}[section]
\newtheorem{proposition}[theorem]{Proposition}
\newtheorem{lemma}[theorem]{Lemma}

\theoremstyle{definition}

\theoremstyle{remark}

\usepackage{url}

\usepackage{newfloat}
\usepackage{listings}
\DeclareCaptionStyle{ruled}{labelfont=normalfont,labelsep=colon,strut=off} 
\floatstyle{ruled}
\newfloat{listing}{tb}{lst}{}
\floatname{listing}{Listing}
\title{Chasing Fairness in Graphs: A GNN Architecture Perspective}

\author{
    Zhimeng Jiang\textsuperscript{\rm 1}, 
    Xiaotian Han\textsuperscript{\rm 1},
    Chao Fan\textsuperscript{\rm 2},
    Zirui Liu\textsuperscript{\rm 3},
    Na Zou\textsuperscript{\rm 4},
    Ali Mostafavi\textsuperscript{\rm 1},
    Xia Hu\textsuperscript{\rm 3}
}
\affiliations{
    \textsuperscript{\rm 1}Texas A\&M University,
    \textsuperscript{\rm 2}Clemson University,
    \textsuperscript{\rm 3}Rice University,
    \textsuperscript{\rm 4}University of Houston,\\
    \{zhimengj,han, amostafavi\}@tamu.edu, cfan@g.clemson.edu, \{zl105, xia.hu\}@rice.edu, nzou2@central.uh.edu
   
}

\usepackage{bibentry}

\begin{document}

\maketitle

\begin{abstract}
There has been significant progress in improving the performance of graph neural networks (GNNs) through enhancements in graph data, model architecture design, and training strategies. For fairness in graphs, recent studies achieve fair representations and predictions through either graph data pre-processing (e.g., node feature masking, and topology rewiring) or fair training strategies (e.g., regularization, adversarial debiasing, and fair contrastive learning). How to achieve fairness in graphs from the model architecture perspective is less explored. More importantly, GNNs exhibit worse fairness performance compared to multilayer perception since their model architecture (i.e., neighbor aggregation) amplifies biases. To this end, we aim to achieve fairness via a new GNN architecture. We propose \textsf{F}air \textsf{M}essage \textsf{P}assing (FMP) designed within a unified optimization framework for GNNs. Notably, FMP \textit{explicitly} renders sensitive attribute usage in \textit{forward propagation} for node classification task using cross-entropy loss without data pre-processing. In FMP, the aggregation is first adopted to utilize neighbors' information and then the bias mitigation step explicitly pushes demographic group node presentation centers together.
In this way, FMP scheme can aggregate useful information from neighbors and mitigate bias to achieve better fairness and prediction tradeoff performance. 
Experiments on node classification tasks demonstrate that the proposed FMP outperforms several baselines in terms of fairness and accuracy on three real-world datasets. The code is available in {\url{https://github.com/zhimengj0326/FMP}}.
\end{abstract}

\section{Introduction}
\label{sect:intro}
Graph neural networks (GNNs) \citep{kipf2017semi,velivckovic2018graph,wu2019simplifying,ling2023graph,han2022g,han2022geometric} are widely adopted in various domains, such as social media mining \citep{hamilton2017inductive}, knowledge graph \citep{hamaguchi2017knowledge} and recommender system \citep{ying2018graph}, due to remarkable performance in learning representations. Graph learning, a topic with growing popularity, aims to learn node representation containing both topological and attribute information in a given graph. Despite the outstanding performance in various tasks, GNNs often inherit or even amplify societal bias from input graph data \citep{dai2021say}.
The biased node representation largely limits the application of GNNs in many high-stake tasks, such as job hunting \citep{mehrabi2021survey} and crime ratio prediction \citep{suresh2019framework}. Hence, bias mitigation that facilitates the research on fair GNNs is in urgent need and we aim to achieve fair prediction for GNNs.

Data, model architecture, and training strategy are the most popular aspects to improve deep learning performance. For fairness in graphs,
many existing works achieving fair prediction in graphs either rely on graph pre-processing (e.g., node feature masking\citep{kose2021fairness}, and topology rewiring \citep{dong2022edits}) or fair training strategies (e.g., regularization \citep{jiang2022generalized}, adversarial debiasing \citep{dai2021say}, or contrastive learning \citep{Zhu:2020vf,zhu2021graph,agarwal2021towards,ling2023learning}). The GNNs architecture perspective to improve fairness in graphs is less explored. More importantly, GNNs are notorious in terms of fairness since GNN aggregation amplifies bias compared to multilayer perception (MLP) \citep{dai2021say}. From the GNNs architecture perspective, message passing is a critical component to improve fairness in graphs.
Therefore, a natural question is raised: 
\begin{center}
    \textit{Can we achieve fairness via fair message passing using vanilla training loss \footnote{The sensitive attributes are not adopted in vanilla training loss. We only consider node classification tasks and vanilla loss is cross-entropy loss in this paper.} without graph pre-processing?}
\end{center}

In this work, we provide a positive answer by designing a fair message-passing scheme guided by a unified optimization framework \footnote{Many aggregations in popular GNNs can be interpreted as gradient descent step for specific optimization problem with specific step size and initialization~\citep{ma2021unified,zhu2021graph}.} for GNNs. The key idea of achieving fair message passing is aggregation first and then conducting bias mitigation via explicitly chasing consistent demographic group representation centers. Specifically,
we first formulate an optimization problem that integrates fairness and smoothness objectives for graph data. Then, we solve the formulated problem via Fenchel conjugate and gradient descent to generate fair and informative representations, where the property of softmax function is adopted to accelerate the gradient calculation over primal variables. We also interpret the optimization problem solver as two main steps (e.g., aggregation first and then debiasing). Finally, we integrate FMP in graph neural networks to achieve fair and accurate prediction for node classification tasks. We demonstrate the superiority of FMP by examining its effectiveness and efficiency on various real-world datasets.

In short, the contributions can be summarized as follows:
\begin{itemize}[leftmargin=0.6cm, itemindent=.0cm, itemsep=0.0cm, topsep=0.0cm]
    \item We demonstrate proof-of-concept that a meticulously crafted GNN architecture can improve fairness for graph data. Our work offers a fresh outlook in comparison to conventional approaches that focus on data pre-processing and fair training strategy design.

    \item We propose FMP to achieve fairness via explicitly incorporating sensitive attribute information in message passing, guided by a unified optimization framework. Additionally, we introduce an acceleration method based on softmax property to reduce gradient computational complexity. 
    \item  The effectiveness and efficiency of FMP are experimentally evaluated on three real-world datasets. The results show that compared to the state-of-the-art, our FMP exhibits a comparable or superior trade-off between prediction performance and fairness with negligibly computation overhead. 
\end{itemize}

\section{Preliminaries}
\subsection{Notations}
We adopt bold upper-case letters to denote matrix such as $\mathbf{X}$, bold lower-case letters such as $\mathbf{x}$ to denote vectors, and calligraphic font such as $\mathcal{X}$ to denote sets. Given a matrix $\mathbf{X}\in\mathbb{R}^{n\times d}$, the $i$-th row and $j$-th column are denoted as $\mathbf{X}_i$ and $\mathbf{X}_{\cdot,j}$, and the element in $i$-th row and $j$-th column is 
$\mathbf{X}_{i,j}$. We use the Frobenius norm, $l_1$ norm of matrix $\mathbf{X}$ as $||\mathbf{X}||_F=\sqrt{\sum_{i,j}\mathbf{X}_{i,j}^2}$ and $||\mathbf{X}||_1=\sum_{ij}|\mathbf{X}_{ij}|$, respectively. Given two matrices $\mathbf{X}, \mathbf{Y}\in\mathbb{R}^{n \times d}$, the inner product is defined as $\langle\mathbf{X}, \mathbf{Y}\rangle=tr(\mathbf{X}^{\top}\mathbf{Y})$, where $tr(\cdot)$ is the trace of a square matrix. $SF(\mathbf{X})$ represents softmax function with a default normalized column dimension. Let $\mathcal{G}=\{\mathcal{V}, \mathcal{E}\}$ be a graph with the node set $\mathcal{V}=\{v_1, \cdots, v_n\}$ and the undirected edge set $\mathcal{E}=\{e_1, \cdots, e_m\}$, where $n, m$ represent the number of node and edge, respectively. The graph structure $\mathcal{G}$ can be represented as an adjacent matrix $\mathbf{A}\in\mathbb{R}^{n\times n}$, where $\mathbf{A}_{ij}=1$ if existing edge between node $v_i$ and node $v_j$. $\mathcal{N}(i)$ denotes the neighbors
of node $v_i$ and $\tilde{\mathcal{N}}(i)=\mathcal{N}(i)\cup \{v_i\}$ denotes the self-inclusive neighbors.
Suppose that each node is associated with a $d$-dimensional feature vector and a (binary) sensitive attribute, the feature for all nodes and sensitive attribute is denoted as $\mathbf{X}_{ori}=\mathbb{R}^{n\times d}$ and $\mathbf{s}\in \{-1, 1\}^{n}$ \footnote{The sensitive attribute $\mathbf{s}$ is not included in node features matrix $\mathbf{X}_{ori}$.}. Define the sensitive attribute incident vector as $\Delta_{\mathbf{s}}= \frac{\bm{1}_{>0}(\mathbf{s})}{||\bm{1}_{>0}(\mathbf{s})||_1} - \frac{\bm{1}_{>0}(-\mathbf{s})}{||\bm{1}_{>0}(-\mathbf{s})||_1}$ to normalize each sensitive attribute group, where $\bm{1}_{>0}(\mathbf{s})$ is an element-wise indicator function.

\subsection{GNNs as Graph Signal Denoising}
A GNN model is usually composed of several stacking GNN layers. Given a graph $\mathcal{G}$ with $N$ nodes, a GNN layer typically contains feature transformation $\mathbf{X}_{trans}=f_{trans}(\mathbf{X}_{ori})$ and aggregation $\mathbf{X}_{agg}=f_{agg}(\mathbf{X}_{trans})$, where $\mathbf{X}_{ori}\in\mathbb{R}^{n\times d_{in}}$, $\mathbf{X}_{trans}, \mathbf{X}_{agg}\in\mathbb{R}^{n\times d_{out}}$ represent the input and output features. The feature transformation operation transforms the node feature dimension, and \emph{feature aggregation}, updates node features based on neighbors' features and graph topology. Recent works \citep{ma2021unified, zhu2021interpreting} have established the connections between many feature aggregation operations $AGG(\cdot)$  in representative GNNs and a graph signal denoising problem with Laplacian regularization, i.e., recovering a clean signal $\mathbf{F}\in\mathbb{R}^{n\times d_{out}}$ from $\mathbf{X}_{trans}$ with the smooth
assumption over graph $\mathcal{G}$. Here, we introduce several popular GNN architectures, including GCN/SGC, GAT, and PPNP/APPNP, as examples to show the connection from the perspective of graph signal denoising. 

\paragraph{GCN/SGC.} Feature aggregation in Graph Convolutional Network (GCN) or Simplifying Graph Convolutional Network (SGC) is given by $\mathbf{X}_{agg}=\tilde{\mathbf{A}}\mathbf{X}_{trans}$, where $\tilde{\mathbf{A}}=\tilde{\mathbf{D}}^{-\frac{1}{2}}\hat{\mathbf{A}}\tilde{\mathbf{D}}^{-\frac{1}{2}}$ is a normalized self-loop adjacency matrix $\hat{\mathbf{A}}=\mathbf{A}+\mathbf{I}$, and $\tilde{\mathbf{D}}$ is degree matrix of $\tilde{\mathbf{A}}$. Recent works \citep{ma2021unified, zhu2021interpreting} provably demonstrate that such feature aggregation can be interpreted as one-step gradient descent to minimize $tr(\mathbf{F}^{\top}\big(\mathbf{I}-\tilde{\mathbf{A}})\mathbf{F}\big)$ with initialization $\mathbf{F}=\mathbf{X}_{trans}$.

\paragraph{GAT.} Feature aggregation in GAT applies the normalized attention coefficient to compute a linear combination of neighbor's features as $\mathbf{X}_{agg, i}=\sum_{j\in\mathcal{N}(i)}\alpha_{ij}\mathbf{X}_{trans, j}$, where $\alpha_{ij}=softmax_j(e_{ij})$, $e_{ij}=\text{LeakyReLU}(\mathbf{X}_{trans, i}^{\top}\mathbf{w}_i+\mathbf{X}_{trans, j}^{\top}\mathbf{w}_j)$, and $\mathbf{w}_i$ and $\mathbf{w}_j$ are learnable column vectors. Prior study \cite{ma2021unified} demonstrates that one-step gradient descent with adaptive stepsize $\frac{1}{\sum_{j\in\tilde{\mathcal{N}}(i)}(c_i+c_j)}$ for the following objective problem:
\be 
\min\limits_{\mathbf{F}}\sum_{i\in\mathcal{V}}||\mathbf{F}_i-\mathbf{X}_{trans, i}||^2_F + \frac{1}{2}\sum_{i\in\mathcal{V}}c_i\sum_{j\in\tilde{\mathcal{N}}(i)}||\mathbf{F}_i-\mathbf{F}_j||_F^2. \nonumber
\ee 
is actually an attention-based feature aggregation, which is equivalent to GAT if $c_i+c_j$ is equivalent to $e_{ij}$, where $c_i$ is a node-dependent coefficient that measures the local smoothness.

\paragraph{PPNP / APPNP.} Feature aggregation in PPNP and APPNP adopt the aggregation rules as $\mathbf{X}_{agg}=\alpha\Big(\mathbf{I}-(1-\alpha)\tilde{\mathbf{A}}\Big)^{-1}\mathbf{X}_{trans}$ and $\mathbf{X}_{agg}^{k+1}=(1-\alpha)\tilde{\mathbf{A}}\mathbf{X}_{agg}^{k}+\alpha\mathbf{X}_{trans}$. It is shown that they are equivalent to the exact solution and one gradient descent step with stepsize $\frac{\alpha}{2}$ to minimize the following objective problem:
\be
\min\limits_{\mathbf{F}}||\mathbf{F}-\mathbf{X}_{trans}||_F^2+(\frac{1}{\alpha}-1)tr\Big(\mathbf{F}^{\top}(\mathbf{I}-\tilde{\mathbf{A}})\mathbf{F}\Big). \nonumber
\ee

\section{Fair Message Passing}
\begin{figure}[t]
\centering
\includegraphics[width=0.95\linewidth]{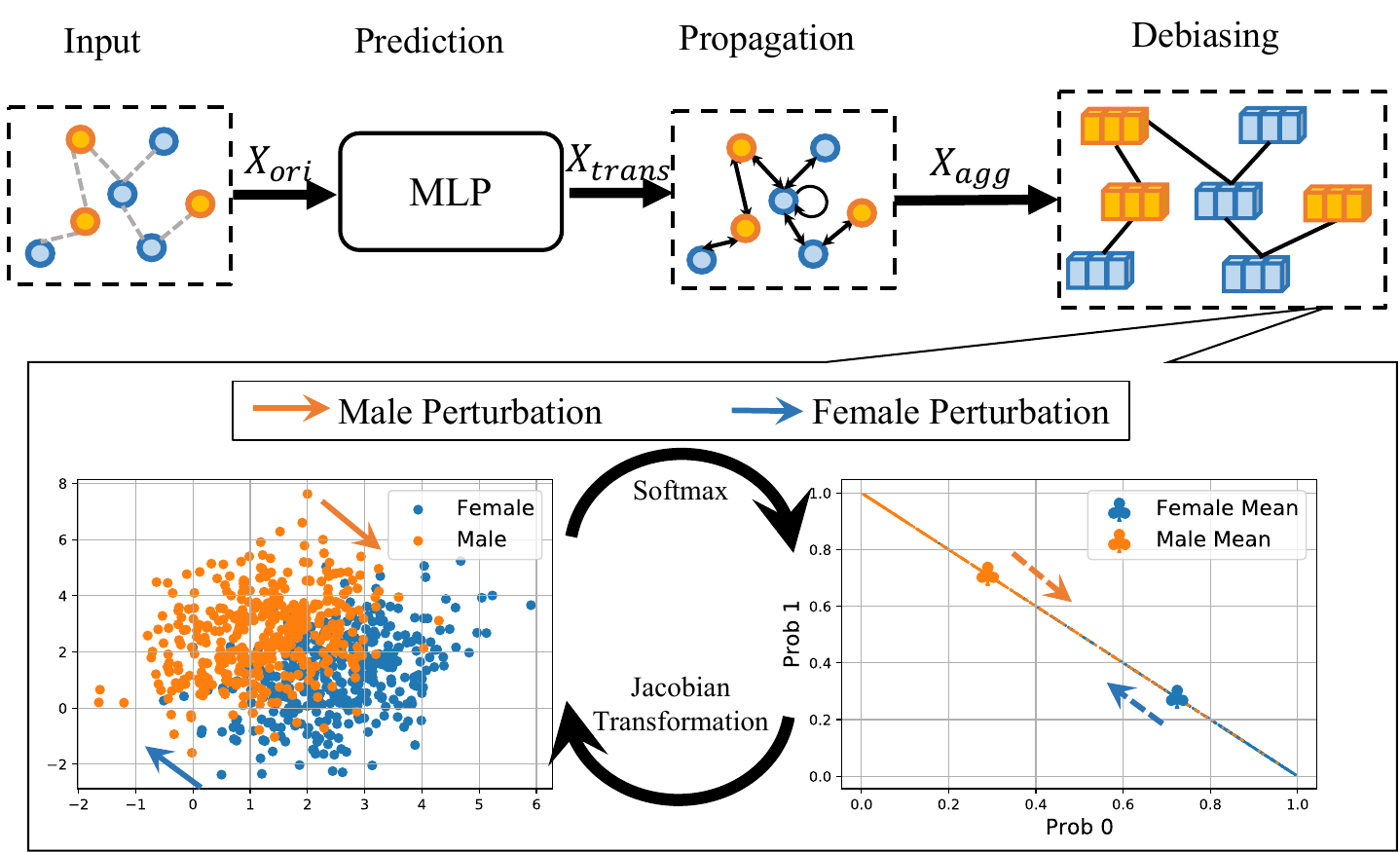}

\caption{The model pipeline consists of three steps: MLP (feature transformation), propagation with skip connection, and debiasing via low-rank perturbation in probability space. }
\vspace{-5pt}
\label{fig:illu}
\end{figure}

In this section, we propose a new fair message-passing scheme to aggregate useful information from neighbors while debiasing representation bias. In this way, fair prediction can be achieved from a model backbone perspective. Specifically, we formulate fair message passing as an optimization problem to pursue \emph{smoothness} and \emph{fair} node representation simultaneously \footnote{Fair message passing is an alternative operation to replace GNNs aggregations.}. Together with an effective and efficient optimization algorithm, we derive the closed-form fair message passing. Finally, the proposed FMP is shown to be integrated into fair GNNs at three stages, including transformation, aggregation, and debiasing step, as shown in Figure~\ref{fig:illu}. These three stages adopted node feature, graph topology, and sensitive attributes respectively.

\subsection{The Optimization Framework}
Most of the existing works rely on hand-craft architecture (e.g., JKNet \citep{xu2018representation}) design for specific tasks, and thus lack of theoretical understanding how such architecture is designed. In our paper, starting from this unified optimization framework for GNNs, we design a new objective, including smoothness and fairness objective, and then derive the proposed FMP to explicitly chase the new objective via fair message passing.

In previous work \citep{ma2021unified}, a general and universal framework is developed to understand aggregation operations in GNNs. Building on top of this framework, we formulate an optimization problem to achieve fair message passing operation (replace aggregation operations in GNNs).
To achieve graph smoothness prior and fairness in the same process, a reasonable message passing should be a good solution for the following optimization problem:
\be \label{eq:optimization}
\min\limits_{\mathbf{F}} \underbrace{\frac{\lambda_{s}}{2}tr(\mathbf{F}^{T}\tilde{\mathbf{L}}\mathbf{F}) + \frac{1}{2} ||\mathbf{F}-\mathbf{X}_{trans}||^2_{F}}_{h_s(\mathbf{F})} + \underbrace{\lambda_{f}||\mathbf{\Delta}_s SF(\mathbf{F})||_{1}}_{h_f\big(\mathbf{\Delta}_s SF(\mathbf{F})\big)}.
\ee  
where $\tilde{\mathbf{L}}$ represents normalized Laplacian matrix, $h_s(\cdot)$ and $h_f(\cdot)$ denotes the smoothness and fairness objectives \footnote{Such smoothness objective is the most common-used one in existing methods \citep{ma2021unified,belkin2001laplacian,kalofolias2016learn}.
The various other smoothness objectives could be considered to improve the performance of FMP and we leave it for future work.}, respectively, and
$\mathbf{X}_{trans}\in \mathbf{R}^{n \times d_{out}}$ is the transformed $d_{out}$-dimensional node features and $\mathbf{F}\in \mathbf{R}^{n \times d_{out}}$ is the aggregated node features of the same matrix size. The first two terms preserve the similarity of connected node representation and thus enforce graph smoothness. The last term enforces fair node representation so that the average predicted probability between groups of different sensitive attributes can remain constant. The regularization coefficients $\lambda_s$ and $\lambda_f$ adaptively control the trade-off between graph smoothness and fairness. 

\paragraph{Smoothness Objective $h_s(\cdot)$.} The adjacent matrix in existing graph message passing schemes is normalized for improving numerical stability and achieving superior performance. Similarly, the graph smoothness term requires normalized Laplacian matrix, i.e., $\tilde{\mathbf{L}}=\mathbf{I}-\tilde{\mathbf{A}}$, $\tilde{\mathbf{A}}=\hat{\mathbf{D}}^{-\frac{1}{2}}\hat{\mathbf{A}}\hat{\mathbf{D}}^{-\frac{1}{2}}$, and $\hat{\mathbf{A}}=\mathbf{A}+\mathbf{I}$. From an edge-centric view, the smoothness objective enforces connected node representation to be similar since 
\be 
tr(\mathbf{F}^{T}\tilde{\mathbf{L}}\mathbf{F})=\sum_{(v_i, v_j)\in\mathcal{E}}||\frac{\mathbf{F}_i}{\sqrt{d_i+1}}-\frac{\mathbf{F}_j}{\sqrt{d_j+1}}||^2_F,
\ee 
where $d_i=\sum_{k}A_{ik}$ represents the degree of node $v_i$.

\paragraph{Fairness Objective $h_f(\cdot)$.} The fairness objective measures the bias for node representation after aggregation. Recall sensitive attribute incident vector $\Delta_{\mathbf{s}}$ indicates the sensitive attribute group and group size via the sign and absolute value summation. Recall that the sensitive attribute incident vector as 
\be 
\Delta_{\mathbf{s}}= \frac{\bm{1}_{>0}(\mathbf{s})}{||\bm{1}_{>0}(\mathbf{s})||_1} - \frac{\bm{1}_{>0}(-\mathbf{s})}{||\bm{1}_{>0}(-\mathbf{s})||_1},
\ee 
and $SF(\mathbf{F})$ represents the predicted probability for node classification task, where $SF(\mathbf{F})_{ij}=\hat{P}(y_i=j|\mathbf{X})$. Furthermore, we can show that our fairness objective is actually equivalent to demographic parity, i.e., $\Big(\Delta_s SF(\mathbf{F})\big)\Big)_j=\hat{P}(y_i=j|\mathbf{s}_i=1, \mathbf{X}) - \hat{P}(y_i=j|\mathbf{s}_i=-1, \mathbf{X})$. Please see proof in \underline{Appendix~\ref{app:fairnessobj}}.
In other words, our fairness objective, $l_1$ norm of $\Delta_s SF(\mathbf{F})$ characterizes the predicted probability difference between two groups with different sensitive attributes. Therefore, our proposed optimization framework can pursue graph smoothness and fairness simultaneously.

\subsection{Optimization Problem Solver}
For smoothness objective, many existing popular message passing schemes can be derived based on gradient descent with appropriate step size choice \citep{ma2021unified,zhu2021interpreting}. In this paper, we consider smoothness objective $h_s(\mathbf{F})$ and fairness objective $h_f(\Delta SF(\mathbf{F}))$ simultaneously for chasing fair and accurate prediction. However, directly solving the optimization problem (\ref{eq:optimization}) is much more challenging due to the nonsmoothness of the fairness objective, and the non-separability of smoothness objective $h_s(\mathbf{F})$ and fairness objective $h_f(\Delta SF(\mathbf{F}))$ due to incident vector $\Delta_s$. 

\subsubsection{Bi-level Optimization Problem Formulation} 
In the literature, many optimization algorithms are developed for optimization problems with $l_1$ norm, such as Alternating Direction Method of Multipliers (ADMM) and Newton type
algorithms~\citep{ghadimi2014optimal,varma2019vector}. However, these algorithms require non-trivial sub-problem solving for each iteration. Therefore, computation complexity is high and is infeasible to integrate deep learning models. Fortunately, Fenchel conjugate (a.k.a. convex conjugate) \citep{rockafellar2015convex} can transform the original problem as an equivalent saddle point problem using a primal-dual
algorithm~\citep{liu2021elastic}. In this way, the computation complexity can be reduced and compatible with back-propagation
training. Similarly, to solve optimization problem \ref{eq:optimization} in a more effective and efficient manner, Fenchel conjugate \citep{rockafellar2015convex} is introduced to transform the original problem 
 into a bi-level optimization problem. For the general convex function $h(\cdot)$, its conjugate function is defined as 
$h^{*}(\mathbf{U})\dff \sup\limits_{\mathbf{X}}\langle \mathbf{U}, \mathbf{X}\rangle -h(\mathbf{X}).$
Based on Fenchel conjugate, the fairness objective can be transformed as variational representation $h_f(\mathbf{p})=\sup\limits_{\mathbf{u}}\langle\mathbf{p},\mathbf{u} \rangle - h_f^{*}(\mathbf{u})$, where $\mathbf{p}=\mathbf{\Delta}_s SF(\mathbf{F})\in\mathbb{R}^{1\times d_{out}}$ is a predicted probability vector for classification. Furthermore, the original optimization problem is equivalent to 
\be \label{eq:minmax}
\min\limits_{\mathbf{F}}\max\limits_{\mathbf{u}} h_s(\mathbf{F}) + \langle\mathbf{p},\mathbf{u} \rangle - h_f^{*}(\mathbf{u})
\ee
where $\mathbf{u}\in\mathbb{R}^{1\times d_{out}}$ and $h_f^{*}(\cdot)$ is the conjugate function of fairness objective $h_f(\cdot)$.

\subsubsection{Problem Solution} 
Motivated by Proximal Alternating Predictor-Corrector (PAPC) \citep{loris2011generalization,chen2013primal}, the min-max optimization problem (\ref{eq:minmax}) can be solved by the following fixed-point equations with per iteration low computation complexity and convergence guarantee
\be 
\left\{ 
\begin{array}{l}
     \mathbf{F}=\mathbf{F}-\nabla h_s(\mathbf{F})-\frac{\partial \langle \mathbf{p}, \mathbf{u}\rangle}{\partial \mathbf{F}}, \\
     \mathbf{u} = \text{prox}_{h^{*}_{f}}\big(\mathbf{u}+ \mathbf{\Delta_{s}} SF(\mathbf{F})\big).
\end{array}
\right.
\ee 
where $\text{prox}_{h^{*}_{f}}(\mathbf{u})=\arg\min\limits_{\mathbf{y}}||\mathbf{y}-\mathbf{u}||_F^2+h^{*}_{f}(\mathbf{y})$. Fortunately, the proximal operators can be obtained with a close form, which makes deep learning model integration feasible. Specifically
we provide the close form of the proximal operators in the following proposition:

\begin{proposition}[Proximal Operators]\label{prop:conjugate}
The proximal operators $\text{prox}_{\beta h^{*}_{f}}(\mathbf{u})$ satisfies 
\be 
\text{prox}_{\beta h^{*}_{f}}(\mathbf{u})_{j}=sign(\mathbf{u})_{j}\min\big(|\mathbf{u}_{j}|, \lambda_f\big),
\ee 
where $sign(\cdot)$ and $\lambda_f$ are element-wise sign function and hyperparameter for fairness objective. In other words, such a proximal operator is an element-wise projection into $l_{\infty}$ ball with radius $\lambda_f$.
\end{proposition}

Similar to ``predictor-corrector" algorithm \citep{loris2011generalization}, we adopt an iterative algorithm to find the saddle point for the min-max optimization problem. Specifically, starting from $(\mathbf{F}^{k}, \mathbf{u}^{k})$, we adopt a gradient descent step on the primal variable $\mathbf{F}$ to arrive $(\bar{\mathbf{F}}^{k+1}, \mathbf{u}^{k})$ and then followed by a proximal ascent step in the dual variable $\mathbf{u}$. Finally, a gradient descent step on a primal variable in point $(\bar{\mathbf{F}}^{k+1}, \mathbf{u}^{k})$ to arrive at $(\mathbf{F}^{k+1}, \mathbf{u}^{k})$. In short, the iteration can be summarized as
\be 
\left\{ 
\begin{array}{l}
     \bar{\mathbf{F}}^{k+1}=\mathbf{F}^{k}-\gamma\nabla h_s(\mathbf{F}^{k})-\gamma\frac{\partial \langle \mathbf{p}, \mathbf{u}^{k}\rangle}{\partial \mathbf{F}}\Big|_{\mathbf{F}^{k}}, \\
     \mathbf{u}^{k+1} = \text{prox}_{\beta h^{*}_{f}}\big(\mathbf{u}^{k}+\beta \mathbf{\Delta_{s}} SF(\bar{\mathbf{F}}^{k+1})\big), \\
     \bar{\mathbf{F}}^{k+1}=\mathbf{F}^{k}-\gamma\nabla h_s(\mathbf{F}^{k})-\gamma\frac{\partial \langle \mathbf{p}, \mathbf{u}^{k+1}\rangle}{\partial \mathbf{F}}\Big|_{\mathbf{F}^{k}}. \\
\end{array}
\right.
\ee 
where $\gamma$ and $\beta$ are the step size for primal and dual variables. Note that the close-form for $\frac{\partial \langle \mathbf{p}, \mathbf{u}\rangle}{\partial \mathbf{F}}\in\mathbb{R}^{n\times d_{out}}$ and $\text{prox}_{\beta h^{*}_{f}}(\cdot)$ are still not clear, we will provide the solution one by one.

\paragraph{FMP Scheme.} Similar to works \citep{ma2021unified,liu2021elastic}, choosing $\gamma=\frac{1}{1+\lambda_s}$ and $\beta=\frac{1}{2\gamma}$, we have
\be 
\mathbf{F}^{k}-\gamma\nabla h_s(\mathbf{F}^{k})&=&\Big((1-\gamma)\mathbf{I}-\gamma\lambda_s\tilde{\mathbf{L}}\Big)\mathbf{F}^{k}+\gamma \mathbf{X}_{trans} \nonumber\\
&=&\gamma \mathbf{X}_{trans} + (1-\gamma)\tilde{\mathbf{A}}\mathbf{F}^{k},
\ee 
Therefore, we can summarize the proposed FMP as two phases, including propagation with skip connection (Step \textbf{\ding{182}}) and bias mitigation (Steps \textbf{\ding{183}}-\textbf{\ding{186}}). For bias mitigation, Step \textbf{\ding{183}} updates the aggregated node features for fairness objective; Steps \textbf{\ding{184}} and \textbf{\ding{185}} aim to learn and ``reshape" perturbation vector in probability space, respectively. Step \textbf{\ding{186}} explicitly mitigates the bias of node features based on gradient descent on the primal variable. The mathematical formulation is given as follows:
\be 
\left\{
\begin{array}{ll}
\mathbf{X}_{agg}^{k+1}=\gamma \mathbf{X}_{trans} + (1-\gamma)\tilde{\mathbf{A}}\mathbf{F}^{k}, & \text{Step \textbf{\ding{182}}}\\
\bar{\mathbf{F}}^{k+1}=\mathbf{X}_{agg}^{k+1}-\gamma \frac{\partial \langle \mathbf{p}, \mathbf{u}^{k}\rangle}{\partial \mathbf{F}}\Big|_{\mathbf{F}^{k}}, & \text{Step \textbf{\ding{183}}}\\
\bar{\mathbf{u}}^{k+1}=\mathbf{u}^{k}+\beta \mathbf{\Delta_{s}} SF(\bar{\mathbf{F}}^{k+1}), & \text{Step \textbf{\ding{184}}}\\
\mathbf{u}^{k+1}=\min\Big(|\bar{\mathbf{u}}^{k+1}|, \lambda_{f} \Big)\cdot sign(\bar{\mathbf{u}}^{k+1}), & \text{Step \textbf{\ding{185}}}\\
\mathbf{F}^{k+1}=\mathbf{X}_{agg}^{k+1}-\gamma \frac{\partial \langle \mathbf{p}, \mathbf{u}^{k+1}\rangle}{\partial \mathbf{F}}\Big|_{\mathbf{F}^{k}}. & \text{Step \textbf{\ding{186}}}
\end{array}
\right. \nonumber
\ee 
where $\mathbf{X}_{agg}^{k+1}$ represents the node features with normal aggregation and skip connection with the transformed input $\mathbf{X}_{trans}$.

\subsubsection{Gradient Computation Acceleration} 
The softmax property is also adopted to accelerate the gradient computation. Note that $\mathbf{p}=\mathbf{\Delta}_s SF(\mathbf{F})$ and $SF(\cdot)$ represents softmax over column dimension, directly computing the gradient $\frac{\partial \langle \mathbf{p}, \mathbf{u}\rangle}{\partial \mathbf{F}}$ based on chain rule involves the three-dimensional tensor $\frac{\partial \mathbf{p}}{\partial \mathbf{F}}$ with gigantic computation complexity. Instead, we simplify the gradient computation based on the property of softmax function in the following theorem. 

\begin{theorem}[Gradient Computation]\label{theo:grad_comp}
The gradient over primal variable $\frac{\partial \langle \mathbf{p}, \mathbf{u}\rangle}{\partial \mathbf{F}}$ satisfies 
\be 
\frac{\partial \langle \mathbf{p}, \mathbf{u}\rangle}{\partial \mathbf{F}}= \mathbf{U}_s\odot SF(\mathbf{F})-\text{Sum}_{1}(\mathbf{U}_s\odot SF(\mathbf{F}))SF(\mathbf{F}).
\ee 
where $\mathbf{U}_s\dff \Delta_s^{\top}\mathbf{u}$, $\odot$ represents the element-wise product and $\text{Sum}_{1}(\cdot)$ represents the summation over column dimension with preserved matrix shape. 
\end{theorem}

\section{Discussion on FMP}
In this section, we provide the interpretation and analyze the \emph{efficiency}, and \emph{white-box usage for sensitive attribute} of the proposed FMP scheme. Furthermore, we also discuss how FMP identifies the influence of sensitive attributes from model forward propagation. 

\paragraph{FMP Interpretation} Note that the gradient of fairness objective over node features $\mathbf{F}$ satisfies $\frac{\partial \langle \mathbf{p}, \mathbf{u}\rangle}{\partial \mathbf{F}}=\frac{\partial \langle \mathbf{p}, \mathbf{u}\rangle}{\partial SF(\mathbf{F})}\frac{\partial SF(\mathbf{F})}{\partial \mathbf{F}}$ and $\frac{\partial \langle \mathbf{p}, \mathbf{u}\rangle}{\partial SF(\mathbf{F})}=\Delta_s^{\top}\mathbf{u}$, such gradient calculation can be interpreted as three steps: Softmax transformation, perturbation in probability space, and debiasing in representation space. Specifically, we first map the node representation into probability space via softmax transformation. Subsequently, we calculate the gradient of fairness objective in probability space. It is seen that the perturbation $\Delta_s^{\top}\mathbf{u}$ actually poses \emph{low-rank} debiasing in probability space, where the nodes with different sensitive attributes embrace opposite perturbations. In other words, \emph{the dual variable $\mathbf{u}$ represents the perturbation direction in probability space.}
Finally, the perturbation in probability space will be transformed into representation space via Jacobian transformation $\frac{\partial SF(\mathbf{F})}{\partial \mathbf{F}}$.

\paragraph{Efficiency.} FMP is an efficient message-passing scheme. The computation complexity for the aggregation (sparse matrix multiplications) is $O(md_{out})$, where $m$ is the number of edges in the graph.
For FMP, the extra computation mainly focuses on the perturbation calculation, as shown in Theorem~\ref{theo:grad_comp}, with the computation complexity $O(nd_{out})$. The extra computation complexity is negligible in that the number of nodes $n$ is far less than the number of edges $m$ in the real-world graph. Additionally, if directly adopting backward propagation to calculate the gradient, we have to calculate the three-dimensional tensor $\frac{\partial \mathbf{p}}{\partial \mathbf{F}}$ with computation complexity $O(n^2d_{out})$. In other words, 
thanks to the softmax property, we achieve an efficient fair message-passing scheme.

\paragraph{White-box Usage for Sensitive Attribute.} The proposed FMP explicitly achieves graph smoothness and fairness objectives via alternative gradient descent. In other words, the usage of sensitive attributes in propagation to mitigate bias is in a white-box manner. Note that such white-box usage of sensitive attributes is a promising property to understand how sensitive attribute usage forces fairness, which is not achieved by previous fairness methods in GNNs. For example, fair training loss utilizes sensitive attributes to regularize the behavior of model prediction and obtain fairer model parameters via rectifying gradients w.r.t. model parameters. In other words, the sensitive attribute information is implicitly encoded in the well-trained model parameters, which makes it hard to understand how sensitive attribute usage helps fair prediction.  Pre-processing fairness methods adopt sensitive attributes to revise data (e.g., node masking and topology rewiring) either in a learnable way or via pre-defined several operations (e.g., node masking and edge deletions). Similarly, the sensitive attribute information is implicitly encoded in the processed data. The understanding of fairness prediction achievement is infeasible. Our FMP can provide a white-box usage for sensitive attributes since we can directly identify that the usage of sensitive attributes is to force the demographic group node representation centers together during forward propagation. 

\begin{table*}[t]

\fontsize{8.0}{14}\selectfont  
\setlength{\tabcolsep}{1.8pt}

\begin{center}
\caption{Comparative Results with Baselines on Node Classification.}
\label{table:comp_gnns}

    \begin{tabular}{ c|ccc|ccc|ccc} 
    \toprule
     \multirow{2}*{Models} & \multicolumn{3}{c}{Pokec-z} & \multicolumn{3}{c}{Pokec-n} & \multicolumn{3}{c}{NBA} \\
    \cline{2-10}
     & Acc ($\%$) $\uparrow$ & $\Delta_{DP}$ ($\%$) $\downarrow$ & $\Delta_{EO}$ ($\%$) $\downarrow$ & Acc ($\%$) $\uparrow$ & $\Delta_{DP}$ ($\%$) $\downarrow$ & $\Delta_{EO}$ ($\%$) $\downarrow$ & Acc ($\%$) $\uparrow$ & $\Delta_{DP}$ ($\%$) $\downarrow$ & $\Delta_{EO}$ ($\%$) $\downarrow$ \\
    \hline
    MLP &  70.48 $\pm$ 0.77 & 1.61 $\pm$ 1.29 & 2.22 $\pm$ 1.01 & 72.48 $\pm$ 0.26 & 1.53 $\pm$ 0.89 & 3.39 $\pm$ 2.37 & 65.56 $\pm$ 1.62 & 22.37 $\pm$ 1.87 & 18.00 $\pm$ 3.52 \\
    \hline
    GAT &  69.76 $\pm$ 1.30 & 2.39 $\pm$ 0.62 & 2.91 $\pm$ 0.97 & 71.00 $\pm$ 0.48 & 3.71 $\pm$ 2.15 & 7.50 $\pm$ 2.88 & 57.78 $\pm$ 10.65 & 20.12 $\pm$ 16.18 & 13.00 $\pm$ 13.37 \\
    \hline
    GCN & \textbf{71.78} $\pm$ 0.37 & 3.25 $\pm$ 2.35 & 2.36 $\pm$ 2.09 & \textbf{73.09} $\pm$ 0.28 & 3.48 $\pm$ 0.47 & 5.16 $\pm$ 1.38 & 61.90 $\pm$ 1.00 & 23.70 $\pm$ 2.74 & 17.50 $\pm$ 2.63 \\
    \hline
    SGC & 71.24 $\pm$ 0.46 & 4.81 $\pm$ 0.30 & 4.79 $\pm$ 2.27 & 71.46 $\pm$ 0.41 & 2.22 $\pm$ 0.29 & 3.85 $\pm$ 1.63 & 63.17 $\pm$ 0.63 & 22.56 $\pm$ 3.94 & 14.33$\pm$ 2.16 \\
    \hline
    APPNP & 66.91 $\pm$ 1.46 & 3.90 $\pm$ 0.69 & 5.71 $\pm$ 1.29 & 69.80 $\pm$ 0.89 & 1.98 $\pm$ 1.30 & 4.01 $\pm$ 2.36 & 63.80 $\pm$ 1.19 & 26.51 $\pm$ 3.33 & 20.00 $\pm$ 4.56 \\
    \hline 
    JKNet & 66.89 $\pm$	3.79 & 1.28 $\pm$0.96 & 1.79 $\pm$ 0.82 & 63.59 $\pm$	6.36 & 1.91	$\pm$ 2.14 & \textbf{0.70} $\pm$ 0.92 & 67.94 $\pm$ 2.73 & 27.80 $\pm$ 8.41 & 20.33 $\pm$ 7.52 \\
    \hline 
    ML1 & 70.42 $\pm$ 0.40 & 2.35	$\pm$ 0.83 & 2.00 $\pm$ 0.50 & 72.36 $\pm$ 0.26 & 1.47 $\pm$ 1.12 &	3.03 $\pm$ 1.77 & 72.70 $\pm$ 1.19 & 26.46 $\pm$ 4.93 &	25.50 $\pm$ 8.38 \\
    \bottomrule
    FMP & 70.50 $\pm$ 0.50 & \textbf{0.81} $\pm$ 0.40 & \textbf{1.73} $\pm$ 1.03 & 72.16 $\pm$ 0.33 & \textbf{0.66} $\pm$ 0.40 & 1.47 $\pm$ 0.87 & \textbf{73.33} $\pm$ 1.85 & \textbf{18.92} $\pm$ 2.28 & \textbf{13.33} $\pm$ 5.89 \\
    \bottomrule
    \end{tabular}
\end{center}
\vspace{-15pt}
\end{table*}

\section{Experiments}
In this section, we conduct experiments to validate the effectiveness and efficiency of the proposed FMP. We firstly validate that graph data with large sensitive homophily enhances bias in GNNs via synthetic experiments. Moreover, for experiments on real-world datasets, we
introduce the experimental settings and then evaluate our proposed FMP compared with several baselines in terms of prediction performance and fairness metrics. 

\subsection{Experimental Settings}
\paragraph{Datasets.} We conduct experiments on real-world datasets Pokec-z, Pokec-n \footnote{Pokec-z and Pockec-n datasets are available at \url{https://github.com/EnyanDai/FairGNN/tree/main}.}, and NBA \citep{dai2021say}. Pokec-z and Pokec-n are sampled, based on province information, from a larger Facebook-like social network Pokec \citep{takac2012data} in Slovakia, where region information is treated as the sensitive attribute and the predicted label is the working field of the users. NBA dataset is extended from a Kaggle dataset \footnote{https://www.kaggle.com/noahgift/social-power-nba} consisting of around 400 NBA basketball players. The information of players includes age, nationality, and salary in the 2016-2017 season. The players' link relationships are from Twitter with the official crawling API. The binary nationality (U.S. and overseas player) is adopted as the sensitive attribute and the prediction label is whether the salary is higher than the median. 

\paragraph{Evaluation Metrics.} We adopt accuracy to evaluate the performance of node classification tasks. As for fairness metrics, we adopt two quantitative group fairness metrics to measure the prediction bias. According to works \citep{louizos2015variational,beutel2017data}, we adopt \emph{demographic parity} $\Delta_{DP}=|\mathbb{P}(\hat{y}=1|s=-1)-\mathbb{P}(\hat{y}=1|s=1)|$ and \emph{equal opportunity} $\Delta_{EO}=|\mathbb{P}(\hat{y}=1|s=-1, y=1)-\mathbb{P}(\hat{y}=1|s=1, y=1)|$, where $y$ and $\hat{y}$ represent the ground-truth label and predicted label, respectively. 

\paragraph{Baselines.} We compare our proposed FMP with representative GNNs, such as GCN \citep{kipf2017semi}, GAT \citep{velivckovic2018graph}, SGC \citep{wu2019simplifying}, and APPNP \citep{klicpera2019predict}, JKNet \citep{xu2018representation}, and MLP. We also compared with method ``ML1" directly using the gradient of Eq. (\ref{eq:optimization}) during model forward propagation. For all models, we train 2 layers of neural networks with 64 hidden units for $300$ epochs. Additionally, We also compare adversarial debiasing and adding demographic regularization methods to show the effectiveness of the proposed method \footnote{Please see the comparison with Fair Mixup \citep{chuang2021fair} in Appendix~\ref{app:fairmixup}}.

\paragraph{Implementation Details.} We run the experiments $5$ times and report the average performance for each method. We adopt Adam optimizer with $0.001$ learning rate and $10^{-5}$ weight decay for all models.
For adversarial debiasing, we adopt the train classifier and adversary with $70$ and $30$ epochs, respectively. 
The hyperparameter for adversary loss is tuned in $\{0.0, 1.0, 2.0, 5.0, 8.0, 10.0, 20.0, 30.0\}$. For adding regularization, we adopt the hyperparameter set $\{0.0, 1.0, 2.0, 5.0, 8.0, 10.0, 20.0, 50.0, 80.0, 100.0\}$.

\begin{figure*}[t]
\centering
\includegraphics[width=0.95\linewidth]{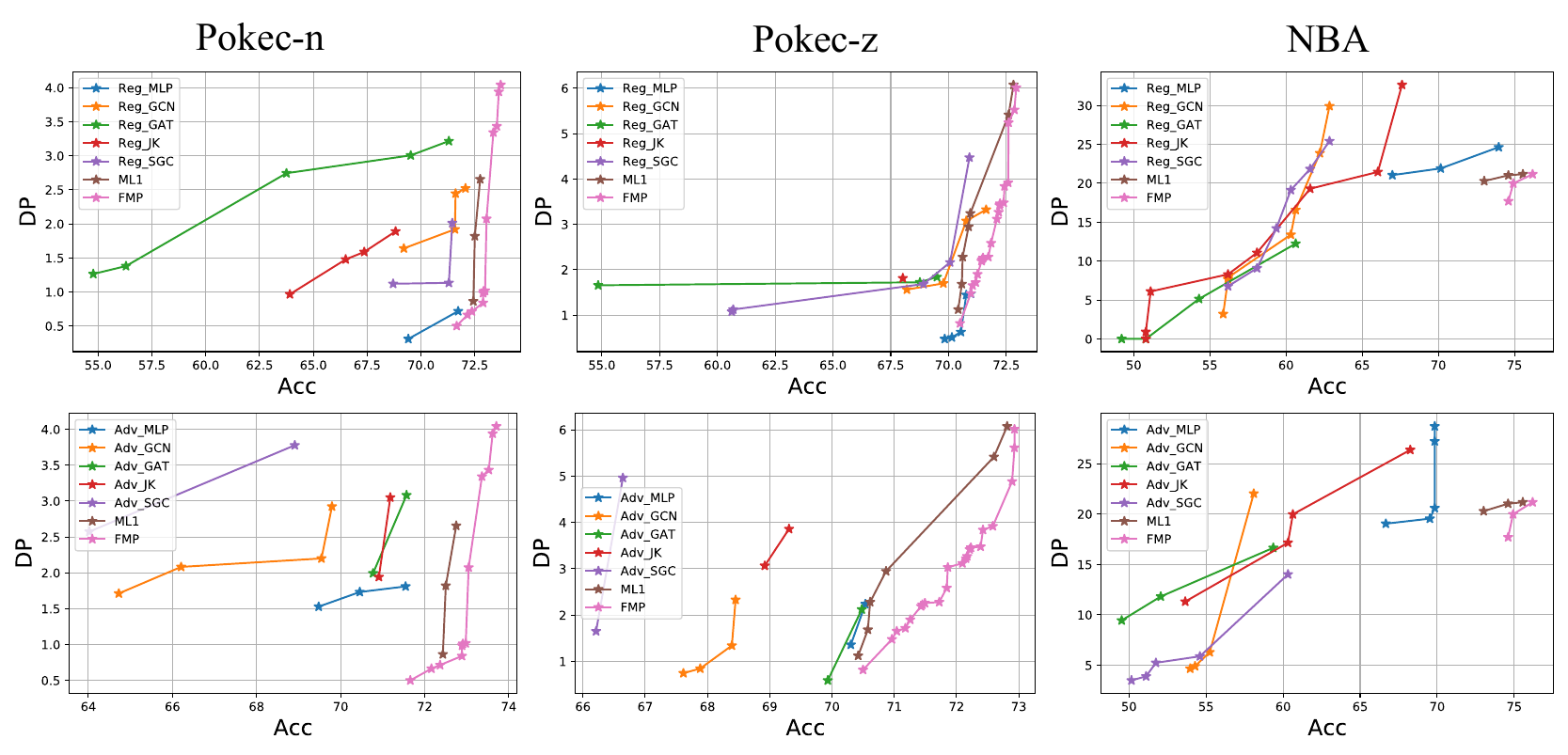}

\caption{DP and Acc trade-off performance on three real-world datasets compared with adding regularization (Top) and adversarial debiasing (Bottom). The trade-off curve close to the right bottom corner means better trade-off performance. The units for x- and y-axis are percentages ($\%$).}
\label{fig:pareto}
\end{figure*}

\subsection{Experimental Results}
\paragraph{Comparison with Existing GNNs.} The accuracy, demographic parity, and equal opportunity metrics of proposed FMP for Pokec-z, Pokec-n, NBA datasets are shown in Table~\ref{table:comp_gnns} compared with MLP, GAT, GCN, SGC, and APPNP. The detailed statistical information for these three datasets is shown in Table ~\ref{table:statistics}. From these results, we can obtain the following observations:
\begin{itemize}[leftmargin=0.2cm, itemindent=.0cm, itemsep=0.0cm, topsep=0.0cm]
    \item Many existing GNNs underperform MLP model on all three datasets in terms of fairness metric. For instance, the demographic parity of MLP is lower than GAT, GCN, SGC and APPNP by $32.64\%$, $50.46\%$, $66.53\%$ and $58.72\%$ on Pokec-z dataset. The higher prediction bias comes from the aggregation within the same sensitive attribute nodes and topology bias in graph data.
    \item Our proposed FMP consistently achieves the lowest prediction bias in terms of demographic parity and equal opportunity on all datasets. Specifically, FMP reduces demographic parity by $49.69\%$, $56.86\%$, and $5.97\%$ compared with the lowest bias among all baselines in Pokec-z, Pokec-n, and NBA datasets. Meanwhile, our proposed FMP achieves the best accuracy in NBA dataset, and comparable accuracy in Pokec-z and Pokec-n datasets. In a nutshell, the proposed FMP can effectively mitigate prediction bias while preserving the prediction performance.
\end{itemize}

\paragraph{Comparison with Adversarial Debiasing and Regularization.} To validate the effectiveness of the proposed FMP, we also show the prediction performance and fairness metric trade-off compared with fairness-boosting methods, including adversarial debiasing \citep{fisher2020debiasing} and adding regularization \citep{chuang2020fair}. Similar to \citep{louppe2017learning}, the output of GNNs is the input of the adversary and the goal of the adversary is to predict the node sensitive attribute. We also adopt several backbones for these two methods, including MLP, GCN, GAT, and SGC. We randomly split $50\%/25\%/25\%$ for training, validation, and test dataset. Figure~\ref{fig:pareto} shows the Pareto optimality curve for all methods, where the right-bottom corner point represents the ideal performance (highest accuracy and lowest prediction bias). From the results, we list the following observations as follows:
\begin{itemize}[leftmargin=0.2cm, itemindent=.0cm, itemsep=0.0cm, topsep=0.0cm]
    \item Our proposed FMP can achieve better DP-Acc trade-off compared with adversarial debiasing and adding regularization for many GNNs and MLP. Such observation validates the effectiveness of the key idea in FMP: aggregation first and then debiasing. Additionally, FMP can reduce demographic parity with negligible performance cost due to transparent and efficient debiasing.
    \item Message passing in GNNs does matter. For adding regularization or adversarial debiasing, different GNNs embrace huge distinctions, which implies that an appropriate message passing manner potentially leads to better trade-off performance. Additionally, many GNNs underperforms MLP in low-label homophily coefficient dataset, such as NBA. The rationale is that aggregation may not always bring benefit in terms of accuracy when the neighbors have low probability with the same label.
\end{itemize}

\section{Related Works} \label{sect:related}
\paragraph{Graph Neural Networks.} GNNs generalizing neural networks for graph data have already shown great success in various real-world applications. There are two streams in GNNs model design, i.e., spectral-based and spatial-based. Spectral-based GNNs provide graph convolution definition based on graph theory, which is utilized in GNN layers together with feature transformation \citep{bruna2013spectral,defferrard2016convolutional,henaff2015deep}. Graph convolutional networks (GCN) \citep{kipf2017semi} simplify spectral-based GNN model into spatial aggregation scheme. Since then, many spatial-based GNNs variant is developed to update node representation via aggregating its neighbors' information, including graph attention network (GAT) \citep{velivckovic2018graph},  GraphSAGE \citep{hamilton2017inductive}, SGC \citep{wu2019simplifying}, APPNP \citep{klicpera2019predict}, et al \citep{gao2018large,monti2017geometric}. Graph signal denoising is another perspective to understand GNNs. Recently, there are several works show that GCN is equivalent to the first-order approximation for graph denoising with Laplacian regularization \citep{henaff2015deep,zhao2019pairnorm}. The unified optimization framework is provided to unify many existing message passing schemes \citep{ma2021unified,zhu2021interpreting}. 

\paragraph{Fairness-aware Learning on Graphs.} Many works have been developed to achieve fairness in machine learning community \citep{jiang2022generalized,han2023retiring,jiang2023weight, chuang2020fair,zhang2018mitigating,du2021fairness,yurochkin2020sensei,creager2019flexibly,feldman2015certifying}. A pilot study on fair node representation learning is developed based on random walk \citep{rahman2019fairwalk}. Additionally, adversarial debiasing is adopted to learn fair prediction or node representation so that the well-trained adversary can not predict the sensitive attribute based on node representation or prediction \citep{dai2021say,bose2019compositional,fisher2020debiasing}. A Bayesian approach is developed to learn fair node representation via encoding sensitive information in the prior distribution in \citep{buyl2020debayes}. Work \citep{ma2021subgroup} develops a PAC-Bayesian analysis to connect subgroup generalization with accuracy parity. \citep{laclau2021all,li2021dyadic} aims to mitigate prediction bias for link prediction. Fairness-aware graph contrastive learning is proposed in \citep{agarwal2021towards,kose2021fairness,ling2023learning}. Graph data preprocessing, such as node feature masking and graph topology rewire, are also developed in \citep{laclau2021all,li2021dyadic,dong2021individual,wang2022improving,zha2023data} for node classification and link prediction tasks. However, the aforementioned works ignore the requirement of transparency in fairness. In this work, we develop an efficient and transparent fair message passing scheme explicitly rendering sensitive attribute usage. 

\section{Conclusion}
In this work, we improve fairness in graphs from the model architecture perspective. We design a fair message-passing scheme to achieve fair prediction for node classification using vanilla training loss without data pre-processing. Specifically,
motivated by the unified optimization framework for GNNs, FMP is designed as aggregation first and then bias mitigation to explicitly chase smoothness and fairness objectives. We also provide a comprehensive discussion of FMP from model architecture interpretation, efficiency, and the white-box usage of sensitive attributes aspects. Experimental results on real-world datasets demonstrate the effectiveness of FMP compared with several baselines in node classification tasks.

\section*{Acknowledgements}
We would like to sincerely thank everyone who has provided their generous feedback for this work. Thank the anonymous reviewers for their thorough comments and suggestions. The work is in part supported by NSF grants
IIS-1900990, IIS-1939716, and IIS-2239257. The views and conclusions
contained in this paper are those of the authors and should
not be interpreted as representing any funding agencies.

\bibliography{fmp}

\clearpage

\newpage
\appendix
\onecolumn
\section{Notations}
\begin{table*}[h]
\caption{Table of Notations}
\centering
    \begin{tabular}{ cc} 
    \toprule
    Notations & Description \\
    \hline
    $|\mathcal{E}|$ & The number of edges \\
    $n$ & The number of nodes\\
    $d$ & The number of node feature dimensions\\
    $d_{out}$ & The number of node classes\\
    $\Delta_{\mathbf{s}}\in\mathbb{R}^{1\times n}$ & The sensitive attribute incident vector \\
    $\epsilon_{label}$ & Label homophily coefficient \\
    $\epsilon_{sens}$ & Sensitive homophily coefficient \\
    $\mathbf{X}_{ori}\in\mathbb{R}^{n\times d}$ & The input node attributes matrix \\
    $\mathbf{A}\in\mathbb{R}^{n\times n}$ & The adjacency matrix\\
    $\hat{\mathbf{A}}\in\mathbb{R}^{n\times n}$ & The adjacency matrix with self-loop\\
    $\tilde{\mathbf{A}}\in\mathbb{R}^{n\times n}$ & The normalized adjacency matrix with self-loop\\
    $\mathbf{L}\in\mathbb{R}^{n\times n}$ & The Laplacian matrix\\
    $\mathbf{X}_{trans}\in\mathbb{R}^{n\times d_{out}}$ & The output node features for feature transformation \\
    $\mathbf{F}_{agg}\in\mathbb{R}^{n\times d_{out}}$ & The aggregated node features after propagation \\
    $\mathbf{F}\in\mathbb{R}^{n\times d_{out}}$ & The learned node features considering graph smoothness and fairness \\
    $\mathbf{u}\in\mathbb{R}^{1\times d_{out}}$ & The permutation direction in feature representation space \\
    $h^{*}(\cdot)$ & Fenchel conjugate function of $h(\cdot)$ \\
    $||\mathbf{X}||_F$, $||\mathbf{X}||_1$ & The Frobenius norm and $l_1$ norm of matrix $\mathbf{X}$\\
    $\lambda_f$, $\lambda_s$ & Hyperparameter for fairness and graph smoothness objectives \\
    \bottomrule 
    \end{tabular}

\end{table*}

\section{Proof on Fairness Objective}\label{app:fairnessobj}
The fairness objective can be shown as the average prediction probability difference as follows:
\be 
\Big(\Delta_s SF(\mathbf{F})\big)\Big)_j&=& \Big[\frac{\bm{1}_{>0}(\mathbf{s})}{||\bm{1}_{>0}(\mathbf{s})||_1} - \frac{\bm{1}_{>0}(-\mathbf{s})}{||\bm{1}_{>0}(-\mathbf{s})||_1}\Big] \big(SF(\mathbf{F})\big)_{:,j}\nonumber\\
&=&\frac{\sum_{\mathbf{s}_i=1}\hat{P}(y_i=j|\mathbf{X})}{||\bm{1}_{>0}(\mathbf{s})||_1}-\frac{\sum_{\mathbf{s}_i=-1}\hat{P}(y_i=j|\mathbf{X})}{||\bm{1}_{>0}(\mathbf{-s})||_1}\nonumber\\
&=&\hat{P}(y_i=j|\mathbf{s}_i=1, \mathbf{X}) - \hat{P}(y_i=j|\mathbf{s}_i=-1, \mathbf{X}). \nonumber
\ee

\section{Proof of Theorem~\ref{theo:grad_comp}}
Before providing in-depth analysis on the gradient computation, we first introduce the softmax function derivative property in the following lemma:
\begin{lemma}\label{lemma:softmax}
For the softmax function with $N$-dimensional vector input $\mathbf{y}=SF(\mathbf{x}):\mathbb{R}^{1\times N} \longrightarrow\mathbb{R}^{1\times N}$, where $y_j=\frac{e^{\mathbf{x}_j}}{\sum_{k=1}^{N}e^{\mathbf{x}_k}}$ for $\forall j\in \{1,2,\cdots, N\}$, the derivative $N\times N$ Jocobian matrix is defined by $[\frac{\partial \mathbf{y}}{\partial \mathbf{x}}]_{ij}=\frac{\partial \mathbf{y}_i}{\partial \mathbf{x}_j}$. Additionally, Jocobian matrix satisfies $\frac{\partial \mathbf{y}}{\partial \mathbf{x}}=\diag(\mathbf{y})-\mathbf{y}^{\top}\mathbf{y}$, where $\mathbf{I}_{N}$ represents $N\times N$ identity matrix and $\top$ denotes the transpose operation for vector or matrix.
\begin{proof}
Considering the gradient $\frac{\partial \mathbf{y}_i}{\partial \mathbf{x}_j}$ for arbitrary $i=j$, according to quotient and chain rule of derivatives, we have
\be 
\frac{\partial \mathbf{y}_i}{\partial \mathbf{x}_j}=\frac{e^{\mathbf{x}_i}\sum_{k=1}^{N}e^{\mathbf{x}_k} -e^{\mathbf{x}_i+\mathbf{x}_j}}{\big(\sum_{k=1}^{N}e^{\mathbf{x}_k}\big)^2}=\frac{e^{\mathbf{x}_i}}{\sum_{k=1}^{N}e^{\mathbf{x}_k}} \cdot \frac{\sum_{k=1}^{N}e^{\mathbf{x}_k}-e^{\mathbf{x}_i}}{\sum_{k=1}^{N}e^{\mathbf{x}_k}}=\mathbf{y}_i (1-\mathbf{y}_j),
\ee 
Similarly, for arbitrary $i\neq j$, the gradient is given by
\be 
\frac{\partial \mathbf{y}_i}{\partial \mathbf{x}_j}=\frac{e^{\mathbf{x}_i}}{\sum_{k=1}^{N}e^{\mathbf{x}_k}} \cdot \frac{-e^{\mathbf{x}_i}}{\sum_{k=1}^{N}e^{\mathbf{x}_k}}=-\mathbf{y}_i\mathbf{y}_j.
\ee 
Combining these two cases, it is easy to verify the Jacobian matrix satisfies $\frac{\partial \mathbf{y}}{\partial \mathbf{x}}=\diag(\mathbf{y})-\mathbf{y}^{\top}\mathbf{y}$.
\end{proof}
\end{lemma}
Arming with the derivative property of softmax function, we further investigate the gradient $\frac{\partial \langle \mathbf{p}, \mathbf{u}\rangle}{\partial \mathbf{F}}$, where $\mathbf{p}=\mathbf{\Delta}_s SF(\mathbf{F})\in \mathbb{R}^{1\times d_{out}}$ and $SF(\cdot)$ and $\mathbf{u}\in \mathbb{R}^{1\times d_{out}}$ is independent with $\mathbf{F}\in \mathbb{R}^{n\times d_{out}}$. 

Considering softmax function $SF(\mathbf{x})\in\mathbb{R}^{n\times d}$ is row-wise adopted in node representation matrix, the gradient satisfies $\frac{\partial SF(\mathbf{F})_{i}}{\partial \mathbf{F}_j}=\mathbf{0}_{d_{out}\times d_{out}}$ for $i\neq j$. Note that the inner product $\langle \mathbf{p}, \mathbf{u}\rangle=\sum_{k=1}^{d_{out}} \mathbf{p}_k \mathbf{u}_k$, it is easy the obtain the gradient $[\frac{\partial \langle \mathbf{p}, \mathbf{u}\rangle}{\partial \mathbf{F}}]_{ij}=\sum_{k=1}^{d_{out}} \frac{\partial \mathbf{p}_k}{\partial \mathbf{F}_{ij}} \mathbf{u}_k$.

To simply the current notation, we denote $\mathbf{\tilde{F}}\dff SF(\mathbf{F})$. According to the chain rule of derivative, we have
\be 
\frac{\partial \mathbf{p}_k}{\partial \mathbf{F}_{ij}} =
\sum_{t=1}^{d_{out}}\frac{\partial \mathbf{p}_k}{\partial \tilde{\mathbf{F}}_{tk}}\frac{\partial \tilde{\mathbf{F}}_{tk}}{\partial \mathbf{F}_{ij}}=
\sum_{t=1}^{d_{out}}\Delta_{\mathbf{s}, t}\frac{\partial \tilde{\mathbf{F}}_{tk}}{\partial \mathbf{F}_{ij}} \overset{(a)}{=}\Delta_{\mathbf{s}, i}\frac{\partial \tilde{\mathbf{F}}_{ik}}{\partial \mathbf{F}_{ij}}\overset{(b)}{=}\Delta_{\mathbf{s}, i}\tilde{\mathbf{F}}_{ik}[\delta_{kj}-\tilde{\mathbf{F}}_{ij}],
\ee 
where $\delta_{kj}$ is Dirac function (equals $1$ only if $k=j$, otherwise $0$;), equality (a) holds since softmax function is row-wise operation, and equality (b) is based on Lemma~\ref{lemma:softmax}. Furthermore, we can obtain the gradient of fairness objective w.r.t. node presentation as follows:
\be 
[\frac{\partial \langle \mathbf{p}, \mathbf{u}\rangle}{\partial \mathbf{F}}]_{ij}=\sum_{k=1}^{d_{out}} \frac{\partial \mathbf{p}_k}{\partial \mathbf{F}_{ij}} \mathbf{u}_k=\sum_{k=1}^{d_{out}}\Delta_{\mathbf{s}, i}\tilde{\mathbf{F}}_{ik}[\delta_{kj}-\tilde{\mathbf{F}}_{ij}]\mathbf{u}_k=\Delta_{\mathbf{s}, i}\tilde{\mathbf{F}}_{ij}\mathbf{u}_j-\Delta_{\mathbf{s}, i}\tilde{\mathbf{F}}_{ij}\sum_{k=1}^{d_{out}}\tilde{\mathbf{F}}_{ik}\mathbf{u}_k.
\ee 
Therefore, the matrix formulation is given by
\be 
\frac{\partial \langle \mathbf{p}, \mathbf{u}\rangle}{\partial \mathbf{F}}= \mathbf{U}_s\odot SF(\mathbf{F})-\text{Sum}_{1}(\mathbf{U}_s\odot SF(\mathbf{F}))SF(\mathbf{F}).
\ee 
where $\mathbf{U}_s\dff \Delta_s^{\top}\mathbf{u}\in\mathbb{R}^{n\times d_{out}}$ and $\text{Sum}_{1}(\cdot)$ represents the summation over column dimension with preserved matrix shape. Therefore, the computation complexity for gradient $\frac{\partial \langle \mathbf{p}, \mathbf{u}\rangle}{\partial \mathbf{F}}$ is $O(n d_{out})$.

\section{Proof of Proposition~\ref{prop:conjugate}}
As for the proximal operators, we provide the close form in the following proposition:

\begin{proposition}[Proximal Operators]
The proximal operators $\text{prox}_{\beta h^{*}_{f}}(\mathbf{u})$ satisfies 
\be 
\text{prox}_{\beta h^{*}_{f}}(\mathbf{u})_{j}=sign(\mathbf{u})_{j}\min\big(|\mathbf{u}_{j}|, \lambda_f\big),
\ee 
where $sign(\cdot)$ and $\lambda_f$ are element-wise sign function and hyperparameter for fairness objective. In other words, such a proximal operator is an element-wise projection into $l_{\infty}$ ball with radius $\lambda_f$.
\end{proposition}

We firstly show the conjugate function for general norm function $f(\mathbf{x})=\lambda ||\mathbf{x}||$, where $\mathbf{x}\in\mathbf{R}^{1\times d_{out}}$. The conjugate function of $f(\mathbf{x})$ satisfies 
\be 
f^{*}(\mathbf{y})=\left\{
\begin{array}{lr}
0, &||\mathbf{x}||_{*}\leq \lambda,\\
+\infty, & ||\mathbf{x}||_{*} > \lambda.
\end{array}
\right.
\ee 
where $||\mathbf{x}||_{*}$ is dual norm of the original norm $||\mathbf{x}||$, defined as $||\mathbf{y}||_{*}=\max\limits_{||\mathbf{x}||\leq 1}\mathbf{y}^{\top}\mathbf{x}$. Considering the conjugate function definition $f^{*}(\mathbf{y})=\max\limits_{\mathbf{x}}\mathbf{y}^{\top}\mathbf{x}-\lambda ||\mathbf{x}||$
the analysis can be divided as the following two cases:

\textbf{\ding{182}} If $||\mathbf{y}||_{*}\leq\lambda$, according to the definition of dual norm, we have $\mathbf{y}^{\top}\mathbf{x}\leq ||\mathbf{x}||||\mathbf{y}||_{*}\leq\lambda ||\mathbf{x}||$ for $\forall ||\mathbf{x}||$, where the equality holds if and only if $||\mathbf{x}||=0$. Hence, it is easy to obtain $f^{*}(\mathbf{y})=\max\limits_{\mathbf{x}}\mathbf{y}^{\top}\mathbf{x}-\lambda ||\mathbf{x}||=0$. 

\textbf{\ding{183}} If $||\mathbf{y}||_{*}>\lambda$, note that the dual norm $||\mathbf{y}||_{*}=\max\limits_{||\mathbf{x}||\leq 1}\mathbf{y}^{\top}\mathbf{x}> \lambda$, there exists variables $\hat{\mathbf{x}}$ so that $||\hat{\mathbf{x}}||\leq 1$ and $\hat{\mathbf{x}}^{\top}\mathbf{y}< \lambda$. Therefore, for any constant $t$, we have $f^{*}(\mathbf{y})\geq \mathbf{y}^{\top}(t\mathbf{x})-\lambda ||t\mathbf{x}||=t(\mathbf{y}^{\top}\mathbf{x}-\lambda ||\mathbf{x}||)\overset{t\rightarrow \infty}{\longrightarrow}\infty$.

Based on the aforementioned two cases, it is easy to get the conjugate function for $l_1$ norm (the dual norm is $l_{\infty}$), i.e., the conjugate function for $h_f(\mathbf{x})=\lambda ||\mathbf{x}||_1$ is given by
\be 
h_{f}^{*}(\mathbf{y})=\left\{
\begin{array}{lr}
0, &||\mathbf{x}||_{\infty}\leq \lambda,\\
+\infty, & ||\mathbf{x}||_{\infty} > \lambda.
\end{array}
\right.
\ee 
Given the conjugate function $h_{f}^{*}(\cdot)$, we further investigate the proximal operators $\text{prox}_{h^{*}_{f}}$. Note that $\text{prox}_{h^{*}_{f}}(\mathbf{u})=\arg\min\limits_{\mathbf{y}}||\mathbf{y}-\mathbf{u}||_F^2+h^{*}_{f}(\mathbf{y})=\arg\min\limits_{||\mathbf{y}||_{\infty}\leq\lambda_{f}}||\mathbf{y}-\mathbf{u}||_F^2=\arg\min\limits_{\substack{\mathbf{y}_j\leq\lambda_{f} \\ \forall j\in[d_{out}]}}\sum_{j=1}^{d_{out}}|\mathbf{y}_j-\mathbf{u}_{j}|^2$, the proximal operator problem can be decomposed as element-wise sub-problem, i.e.,
\be 
\text{prox}_{h^{*}_{f}}(\mathbf{u})_{j}=\arg\min\limits_{\mathbf{y}_j\leq\lambda_{f}}|\mathbf{y}_j-\mathbf{u}_{j}|^2=sign(\mathbf{u}_j) \min(|\mathbf{u}_j|, \lambda_{f}) \nonumber
\ee  
which completes the proof.

\section{More discussion on FMP}
To facilitate the understanding of the influence of sensitive attributes, we measure the influence of sensitive attributes as the difference of final prediction between the well-trained fair model using sensitive attributes and vanilla models without sensitive attribute usage. The sensitive attribute has a critical influence to achieve fair prediction and the prediction is highly different for the vanilla model (trained with vanilla loss and no data preprocessing) and the fair model (trained with fair methods). We visualize the logit layer node representation for different methods in Appendix~\ref{app:influprob}. 

The proposed FMP explicitly uses the sensitive attribute information in Steps \textbf{\ding{183}}-\textbf{\ding{186}} during forward propagation. In other words, if we aim to identify the influence of sensitive attributes for FMP, it is sufficient to check the difference between the input and output for the debiasing step since it is disentangled with feature transformation and aggregation. It is worth mentioning that the required information for identifying the influence of sensitive attributes is naturally from the forward propagation. However, for the fair model from existing works (e.g, adding regularization and adversarial debiasing), note that the sensitive attribute information is implicitly encoded in the well-trained model weight, the sensitive attribute perturbation inevitably leads to the variability of well-trained model weight. Therefore, it is required to retrain the model for probing the influence of sensitive attribute perturbation. The key drawback of these methods is due to encoding the sensitive attributes information into well-trained model weights. From the auditors' perspective, it is quite hard to identify the influence of sensitive attributes only given a well-trained fair model. Instead, our designed FMP explicitly adopts the sensitive attribute information in the forward propagation process, which naturally avoid the dilemma that sensitive attributes are encoded into well-trained model weight. In a nutshell, FMP encompasses higher transparency since (1) the sensitive attribute is explicitly adopted in forward propagation; (2) It is not necessary to retrain the model for probing the influence of the sensitive attribute. 

We aim to provide a more precise statement on transparency in fairness (TIF) and then point out why many fair methods can not achieve transparency in fairness. Intuitively, TIF represents that the influence of sensitive attribute in the inference stage for a fair method can be obtained with only a well-trained fair model and test data. Although many fair methods relying on sensitive attribute are developed to achieve a fair model, the process of how the sensitive attribute makes the model to be fair is still black-box. To this end, we introduce TIF, a general concept beyond graph data. Denote training dataset $\mathcal{D}_{train}=\{X_{train}, s_{train}, y_{train}\}$ and test dataset $\mathcal{D}_{test}=\{X_{test}, s_{test}, y_{test}\}$, where $X_{train}$ ($X_{test}$), $s_{train}$ ($s_{test}$), and $y_{train}$ ($y_{test}$) represent the input attributes, sensitive attributes, and label for model training (test).  We first provide a formal statement on the influence of sensitive attribute and TIF for a specific fair method.
\paragraph{What is the influence of sensitive attributes in the inference stage?} The influence of sensitive attributes can be regarded as the difference between the well-trained fair and vanilla model. The fair model $f_{\theta^*}(\cdot)$ can be obtained using training dataset (including sensitive attribute) and a specific fair method (e.g., fair regularization, adversarial debiasing) while vanilla model $f_{\theta_0}(\cdot)$ is obtained without any usage of sensitive attribute (e.g., vanilla loss and no data pre-processing and post-processing). Define $M(f_{\theta}, \mathcal{D}_{test})$ as the measurement (not necessarily scaler) for a well-trained model $f_{\theta}(\cdot)$ given test dataset $\mathcal{D}_{test}$. For example, test loss or model prediction can be instantiations as measurements. Then the influence of sensitive attributes represents the measurement difference between the well-trained fair and vanilla models $M(f_{\theta^*}, \mathcal{D}_{test})-M(f_{\theta_0}, \mathcal{D}_{test})$.

\paragraph{What is TIF?} TIF represents that the influence of sensitive attributes can be obtained via the well-trained fair model and test data (without access to the training data). To obtain the influence of sensitive attribute, the fair model and vanilla model are both required to obtain the influence of sensitive attribute. In other words, for existing fair methods (e.g., pre-processing, in-processing, and post-processing methods), it is intractable to obtain such influence if only having access to the fair model since the training data or vanilla model can not be accessed.

\paragraph{Difference with model interpretability.}   Model interpretability aims to understand and explain the steps and decisions of the model when making predictions. There are two types of interpretability, named intrinsical interpretability and post-hoc interpretability. Intrinsically interpretable models (such as decision trees) can provide human-understandable decision-making from the model itself, while post-hoc interpretability requires external methods to help humans understand how the model makes predictions. Similar to intrinsical interpretability, whether the fair model with TIF is essentially binary. The key difference is that TIF aims to understand how sensitive attribute helps to achieve a fair model for a specific fair method. Such a fairness-achieving process is essentially dynamic while model interpretability is static for model prediction.

\paragraph{The main idea to achieve TIF.} Note that many existing methods, including pre-processing, in-processing, and post-processing methods, can not achieve TIF, we try to integrate sensitive attribute information into the forward propagation for model prediction. In this way, the influence of sensitive attributes can be obtained through model inference. Thanks to the unified optimization framework for GNNs, we develop a fair message passing (FMP), which explicitly and separately uses sensitive attributes in the (last) debiasing stage of forward propagation. which makes it easy to identify the influence. In this way,  the influence of sensitive attribute can be identified using the input and output of the debiasing stage. 

\section{More Details on Computational Complexity of FMP}
The aggregation is conducted using Step \textbf{\ding{182}}. The computation complexity for skip connection and sparse matrix multiplications in aggregation are $O(nd)$ and $O(md)$, respectively, where $m$ and $n$ represent the number of edges and nodes. As for perturbation calculation (Steps \textbf{\ding{183}}-\textbf{\ding{184}}), the computation complexity for Step \textbf{\ding{183}}, Step \textbf{\ding{184}} and Step \textbf{\ding{186}} are all $O(nd)$ based on Eq. (9). The computation for Step \textbf{\ding{185}} is $O(d)$ for sign and min operations. Additionally, if we directly calculate gradient $\frac{\partial \langle \mathbf{p}, \mathbf{u}\rangle}{\partial \mathbf{F}}$ using chain rule, the calculation three-dimensional tensor $\frac{\partial \mathbf{p}}{\partial \mathbf{F}}\in \mathbf{R}^{n\times d^2}$ is involved with computation complexity $O(nd_{out}^2)$. We will revise the computation complexity accordingly.

\section{Training Algorithms}
We summarize the training algorithm for FMP and provide the pseudo codes in Algorithm \ref{algo:FMP}.

\begin{algorithm}[htb]
   \caption{FMP Training Algorithm}
   \label{algo:FMP}
\begin{algorithmic}
   \STATE {\bfseries Input:} Graph dataset $\mathcal=(\mathbf{X}, \mathbf{A}, \mathbf{Y})$; The total epochs $T$; Hyperparameters $\lambda_{s}$ and $\lambda_{f}$.
   \STATE {\bfseries Output:} The well-trained FMP model.
   \STATE Initialize model parameters.
   \FOR{epoch from $1$ to $T$}
   \STATE Conduct feature transformation using MLP
   \STATE Conduct propagation and debiasing as steps \textbf{\ding{182}}-\textbf{\ding{186}}
    \STATE Calculate the cross entropy loss for node classification task
    \STATE Conduct backpropagation step to update model weight
   \ENDFOR
\end{algorithmic}
\end{algorithm}

    

\section{Dataset Statistics}\label{app:stat}
For a fair comparison with previous work, we perform the node classification task on three real-world datasets, including Pokec-n, Pokec-z, and NBA.
The data statistical information on three real-world datasets is provided in Table~\ref{table:statistics}. It is seen that the sensitive homophily are even higher than the label homophily coefficient among three real-world datasets, which validates that the real-world datasets is usually with large topology bias. 

\begin{table*}[h]
\fontsize{12}{12}\selectfont  
\setlength{\tabcolsep}{3pt}
\begin{center}
\caption{Statistical Information on Datasets}
\label{table:statistics}
\scalebox{0.92}{
    \begin{tabular}{ c|c|c|c|c|c} 
    \toprule
    Dataset & $\#$ Nodes & \makecell*[c]{$\#$ Node Features} & $\#$ Edges & $\#$ Training Labels & $\#$ Training Sens \\
    \hline
    Pokec-n &  66569 & 265 & 1034094 & 4398 & 500 \\
    \hline
    Pokec-z & 67796 & 276 & 1235916 & 5131 & 500 \\
    \hline
    NBA & 403 & 95 & 21242 & 156 & 246 \\
    \bottomrule
    \end{tabular}
}
\end{center}
\end{table*}




\section{More Experimental Results}\label{app:more_exp}

\subsection{More Experimental Setting Details}\label{app:exp_detail}
In FMP implementation, we first use 2 layers of MLP with 64 hidden units and the output dimension for MLP is 2. We also stack 2 layers for propagation and debiasing steps, where there are not any trainable model parameters. As for the model training, we adopt cross-entropy loss function with 300 epochs. We also adopt Adam optimizer with $0.001$ learning rate and $1\times 10^{-5}$
weight decay for all models. The hyperprameters for FMP is $\lambda_{f}=\{0, 5, 10, 15, 20, 30, 100\}$ and $\lambda_{s}=\{0, 0.01, 0.1, 0.5, 1.0, 2.0, 3, 5, 10, 15, 20\}$. For adding regularization, we select the hyperparameter set as $\{0.0, 2.0, 5.0, 8.0, 10.0, 15.0, 20.0, 30.0, 50.0, 80.0, 100.0\}$ for all datasets. As for adversarial debiasing, we adopt hyperparameters in $\{0.0, 1.0, 5.0, 8.0, 10.0, 20.0, 30.0, 50.0, 80.0\}$.

\subsection{Comparison with Fair Mixup}\label{app:fairmixup}

\begin{figure*}[t]
\centering
\includegraphics[width=0.99\linewidth]{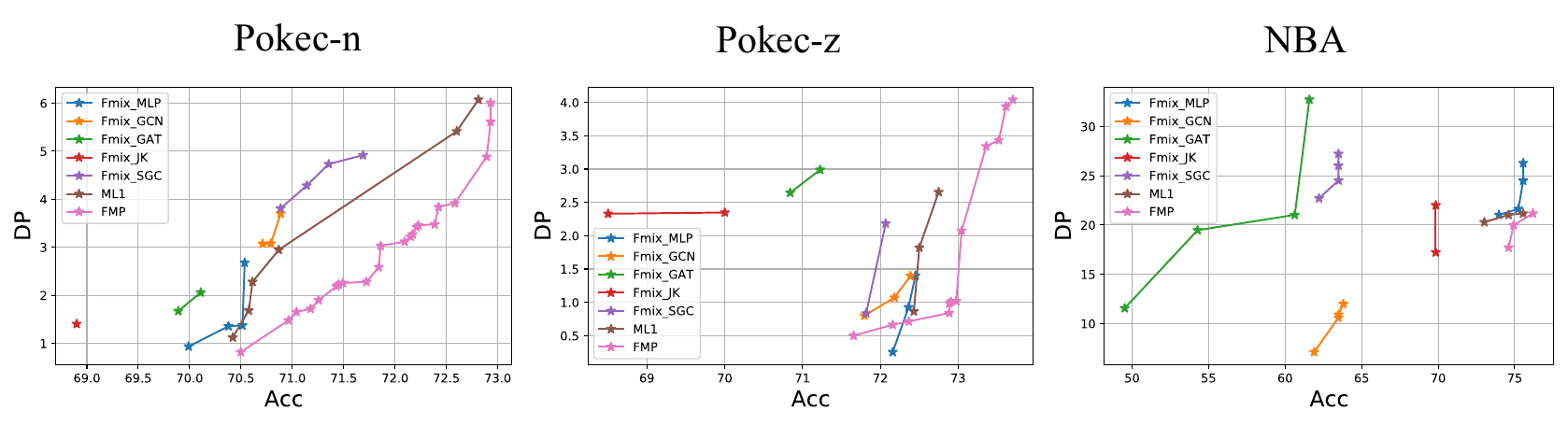}

\caption{DP and Acc trade-off performance on three real-world datasets compared with (manifold) Fair Mixup.}
\label{fig:tradeoff_fairmixup}
\vspace{-5pt}
\end{figure*}

We also implement Fair mixup \citep{chuang2021fair} as the additional baseline for different GNN backbones in Figure~\ref{fig:tradeoff_fairmixup}. Note that input fair mixup requires calculating model prediction for mixed input batch, it is non-trivial to adopt input fair mixup in our experiments (node classification task) since forward propagation in GNN  aggregates information from neighborhoods while the neighborhood information for the mixed input batch is missing. Therefore, we adopt manifold fair mixup for the logit layer (the previous layers contain aggregation step) in our experiments. Experimental results show that our method can still achieve better accuracy-fairness tradeoff performance on three datasets. 

\subsection{Sensitive Attribute Influence Probe}\label{app:influprob}
As for lending fairness perceptron, it represents the influence of sensitive attributes that could be identified. For example, our proposed FMP includes three steps, i.e., transformation, aggregation, and debiasing, where the sensitive attribute is explicitly adopted in debiasing step. If we aim to identify the influence of sensitive attributes for FMP, it is sufficient to check the difference between the input and output for the debiasing step. It is worth mentioning that the required information for identifying the influence of sensitive attributes is naturally from the forward propagation. Additionally, if we aim to identify the influence of sensitive attributes for existing methods (e.g, adding regularization and adversarial debiasing), the well-trained fair model is insufficient and we need additional vanilla (unfair) model without using any sensitive attribute information. In other words, these methods require model retraining with sensitive attribute movement, and thus much more resources for sensitive attributes influence auditing. The key drawback of these methods is due to encoding the sensitive attributes information into well-trained model weights. From the auditors' perspective, it is quite hard to identify the influence of sensitive attributes only given a well-trained fair model. Instead, our designed FMP explicitly adopts the sensitive attribute information in the forward propagation process, which naturally avoid the dilemma that sensitive attributes are encoded into well-trained model weight.

\begin{figure*}[t]
\centering
\includegraphics[width=0.99\linewidth]{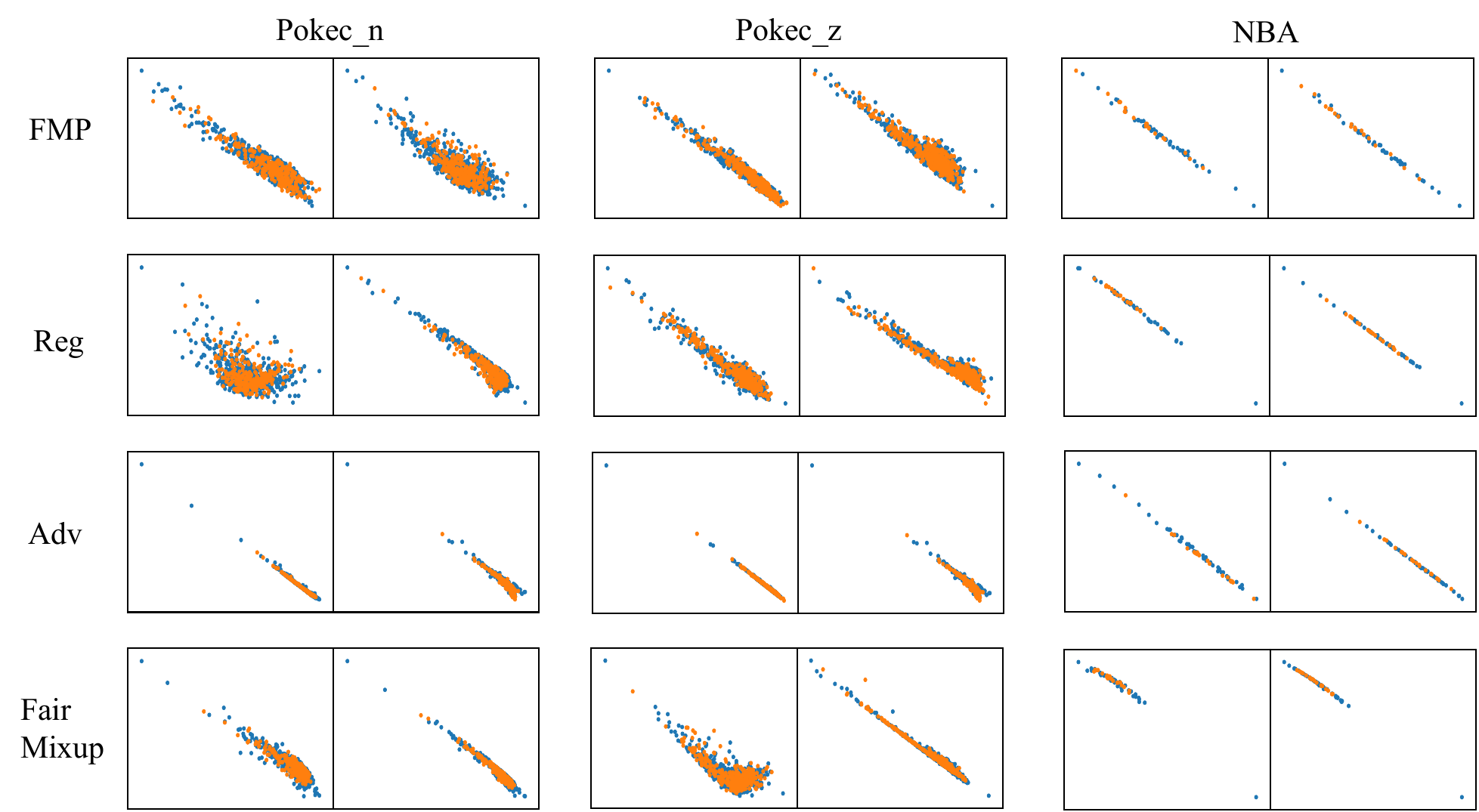}

\caption{The visualization of logit layer node representation for training with/without (left/right) sensitive attribute for FMP and several baselines across three real-world datasets. The data point with different colors represents different sensitive attributes.}
\label{fig:infprobe}
\vspace{-5pt}
\end{figure*}

Figure~\ref{fig:infprobe} shows the visualization results for training with/without (left/right) sensitive attributes for FMP and several baselines (with GCN backbones) across three real-world datasets. From the visualization results, we observe that all methods with sensitive attribute information achieve better fairness since the logit layer representation for different sensitive attributes is mixed with each other. Therefore, it is hard to identify the sensitive attribute based on the representation and thus leads to higher fairness results. The key difference is that the results for training with/without (left/right) sensitive attribute in FMP can both be obtained through forward propagation, while the other baseline methods require model retraining to probe the influence of sensitive attributes.

\subsection{Running Time Comparison} \label{app:runningtime}
\begin{figure*}[t]
\centering
\includegraphics[width=0.99\linewidth]{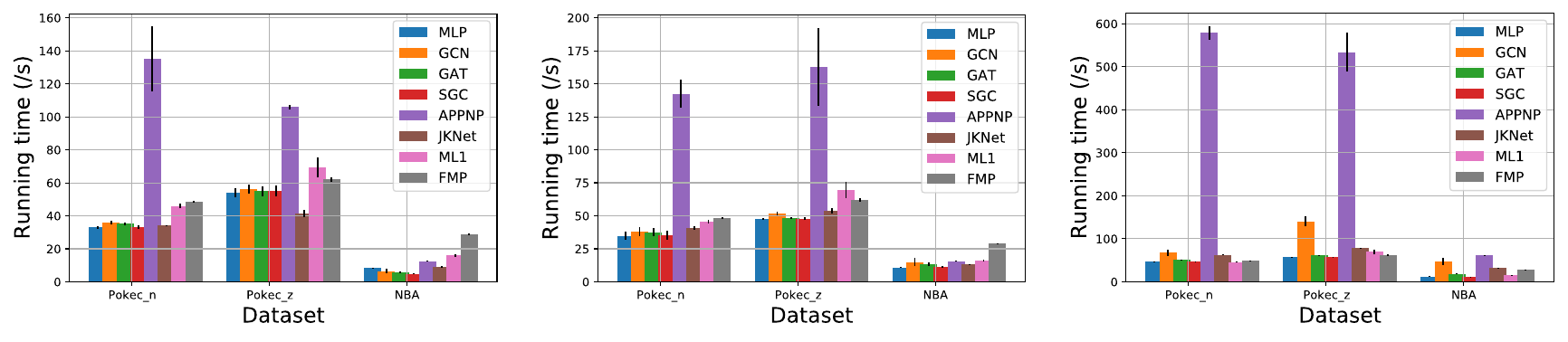}

\caption{The running time comparison.}
\label{fig:runtime}
\vspace{-5pt}
\end{figure*}

We provide running time comparison in Figure~\ref{fig:runtime} for our proposed FMP and other baselines, including vanilla, regularization, and adversarial debiasing on many backbones (MLP, GCN, GAT, SGC, and APPNP). To achieve a fair comparison, we adopt the same Adam optimizer with $200$ epochs with $5$ running times. 
We list several observations as follows:
\begin{itemize}[leftmargin=0.2cm, itemindent=.0cm, itemsep=0.0cm, topsep=0.0cm]
    \item The running time of proposed FMP is very efficient for large-scale datasets. Specifically, for the vanilla method, the running time of FMP is higher than most lighten backbone MLP with $46.97\%$ and $15.03\%$ time overhead on Pokec-n and Poken-z datasets, respectively. Compared with the most time-consumption APPNP, the running time of FMP is lower with $64.07\%$ and $41.45\%$ time overhead on Pokec-n and Poken-z datasets, respectively.
    \item The regularization method achieves almost the same running time compared with the vanilla method on all backbones. For example, GCN with regularization encompasses higher running time with $6.41\%$ time overhead compared with the vanilla method. Adversarial debiasing is extremely time-consuming. For example, GCN with adversarial debiasing encompasses higher running time with $88.58\%$ time overhead compared with the vanilla method.
\end{itemize}

\subsection{Hyperparameter Study} \label{app:hyper}
\begin{figure}[t]
\centering
\includegraphics[width=0.9\linewidth]{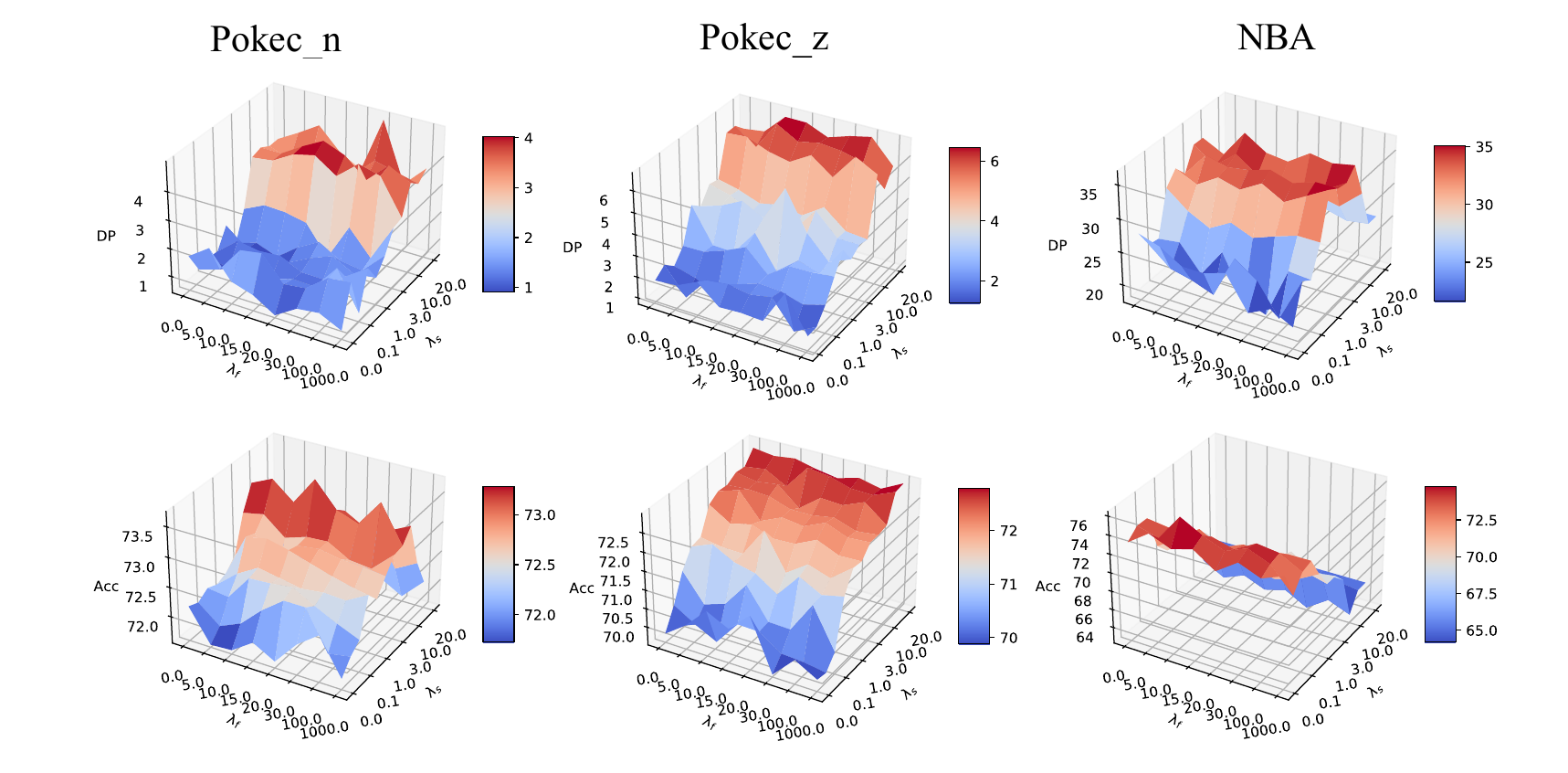}

\caption{Hyperparameter study on fairness and smoothness hyperparameter for demographic parity and Accuracy.}
\label{fig:hyper}
\end{figure}

We provide hyperparameter study for further investigation on fairness and smoothness hyperparmeter on prediction and fairness performance on three datasets. Specifically, we tune hyperparameters as $\lambda_f=\{0.0, 5.0, 10.0, 15.0, 20.0, 30.0, 100.0, 1000.0\}$ and $\lambda_s=\{0.0, 0.1, 0.5, 1.0, 3.0, 5.0, 10.0, 15.0, 20.0\}$. From the results in Figure~\ref{fig:hyper}, we can make the following observations:
\begin{itemize}
    \item The accuracy and demographic parity are extremely sensitive to the smoothness hyperparameter. It is seen that, for Pokec-n and Pokec-z datasets (NBA), a larger smoothness hyperparameter usually leads to higher (lower) accuracy with higher prediction bias. The rationale is that, only for graph data with a high label homophily coefficient, GCN-like aggregation with skip connection is beneficial. Otherwise, the neighbor's node representation with a different label will mislead the representation update.
    \item The appropriate fairness hyperparameter leads to better fairness and prediction performance tradeoff. The reason is that fairness hyperparameter determines the perturbation vector update step size in probability space. Only appropriate step size can lead to better perturbation vector update.
\end{itemize}

\subsection{Results on Additional Datasets}

\begin{figure*}[ht]
\centering
\includegraphics[width=0.8\linewidth]{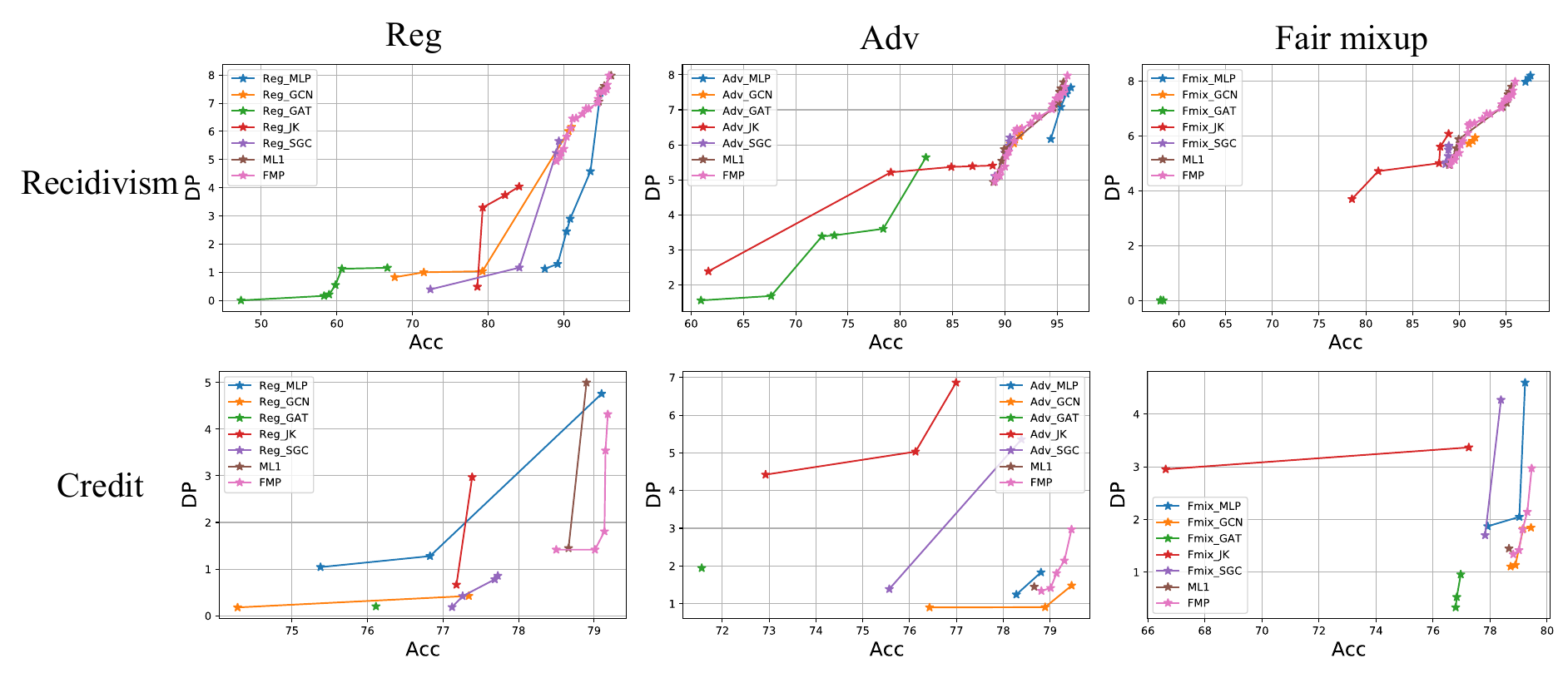}

\caption{DP and Acc trade-off performance on three real-world datasets compared with adding regularization, adversarial debiasing, and (manifold) Fair Mixup in additional datasets.}
\label{fig:tradeoff_manual}
\vspace{-5pt}
\end{figure*}

We also conduct experiments on two new datasets (Recidivism and Credit), where the graph topology is constructed based on node features.  In Recidivism, nodes are defendants
released on bail from 1990 to 2009, where the nodes are connected based on
the similarity of past criminal records and demographics. The task
is to predict defendant is on bail or not, and the
sensitive attribute is selected as ``race". In the Credit dataset, credit card
users (nodes) are connected based on the pattern similarity of
their purchases and payments. The sensitive attribute is selected as ``age", 
and the task is to predict whether a user will default on credit card payment. 
Figure.~\ref{fig:tradeoff_manual} demonstrates the tradeoff performance for different fair methods, including adding regularization, adversarial debiasing, and fair mixup.
Experimental results show that our method can still achieve good accuracy-fairness tradeoff performance on three datasets. We also notice that MLP can achieve good tradeoff performance since the graph topology is manually constructed based on node attribute similarity.

\section{Future Work}
There are three lines of follow-up research directions. Firstly, achieving transparency can be further developed. For example, for the intransparent model, how can we develop external methods to probe the influence of sensitive attributes in the target model? Secondly, given the influence of sensitive attributes, how can we interpret the influence of sensitive attributes in a human-understandable way? For example, how can we measure the benefit of such influence toward fairness? Thirdly, it is also interesting to extend FMP into more general cases, such as continuous sensitive attributes \citep{jiang2022generalized}, and limited sensitive attributes \citep{dai2021say}.

\section{Broader Social Impact and Limitations}\label{app:impact}
Transparency in fairness is an advanced property in the fairness domain and poses huge challenges for research and industry. Many existing works mainly rely on specific fairness metrics to evaluate the prediction bias. Transparency may stimulate maintainers and auditors of machine learning systems to rethink fairness evaluation/auditing. Only achieving a fair model with a lower bias for specific fairness metrics is insufficient. The maintainers should also consider how to leverage the influence of sensitive attributes for auditors. Transparency may lead maintainers to pay more effects to improve the transparency of the fair model and could be helpful to convince the auditors. The limitations of this work are that it requests sensitive information in the inference stage.

\end{document}

%% file: math_commands.tex
\usepackage{amsmath,amsfonts,bm}

\def\eqref#1{equation~\ref{#1}}

\def\1{\bm{1}}

\DeclareMathAlphabet{\mathsfit}{\encodingdefault}{\sfdefault}{m}{sl}
\SetMathAlphabet{\mathsfit}{bold}{\encodingdefault}{\sfdefault}{bx}{n}



%% file: pream_bm.tex
\newtheorem{prop}{Proposition}

\newtheorem{cor}{Corollary}

\newtheorem{lm}{Lemma}

\newtheorem{thm}{Theorem}

\newcommand{\be}{\begin{eqnarray}}
\newcommand{\ee}{\end{eqnarray}}
\newcommand{\benn}{\begin{eqnarray*}}
\newcommand{\eenn}{\end{eqnarray*}}
\def\IR{\rm I \kern-0.20em R}

\newcommand{\bthm}{\begin{thm}}
\newcommand{\ethm}{\end{thm}}

\newcommand{\bcor}{\begin{cor}}
\newcommand{\ecor}{\end{cor}}
\newcommand{\bprop}{\begin{prop}}
\newcommand{\eprop}{\end{prop}}
\newcommand{\blm}{\begin{lm}}
\newcommand{\elm}{\end{lm}}
\newcommand{\beq}{\begin{equation}}
\newcommand{\eeq}{\end{equation}}
\newcommand{\ber}{\begin{eqnarray}}
\newcommand{\eer}{\end{eqnarray}}

\newcommand{\bproof}{\begin{proof}}
\newcommand{\eproof}{\end{proof}}

\newcommand{\diag}{\mathop{\mbox{\rm diag}}}

\newcommand{\bit}{\begin{itemize}}
\newcommand{\eit}{\end{itemize}}
\newcommand{\ben}{\begin{enumerate}}
\newcommand{\een}{\end{enumerate}}
\newcommand{\bdesc}{\begin{description}}
\newcommand{\edesc}{\end{description}}
\newcommand{\beqarrn}{\begin{eqnarray*}}
\newcommand{\eeqarrn}{\end{eqnarray*}}
\newcommand{\bproofof}{\begin{proofof}}
\newcommand{\eproofof}{\end{proofof}}
\newenvironment{rem}{\begin{trivlist}\item[]{\bf
Remark:}\hspace{4mm}}{\end{trivlist}}
\newcommand{\brem}{\begin{rem}}
\newcommand{\erem}{\end{rem}}
\newenvironment{rems}{\begin{trivlist}\item[]{\bf
Remarks}\begin{itemize}}{\end{itemize}\end{trivlist}}
\newcommand{\brems}{\begin{rems}}
\newcommand{\erems}{\end{rems}}
\newtheorem{fact}{Fact}
\newcommand{\bfact}{\begin{fact}}
\newcommand{\efact}{\end{fact}}
\newtheorem{examp}{Example}
\newcommand{\bexamp}{\begin{examp}\rm}
\newcommand{\eexamp}{\end{examp}}
\newtheorem{defn}{Definition}
\newcommand{\bdefn}{\begin{defn}\rm}
\newcommand{\edefn}{\end{defn}}

\newtheorem{alg}{Algorithm}
\newcommand{\balg}{\begin{alg}}
\newcommand{\ealg}{\end{alg}}

\newtheorem{prob}{Problem}
\newcommand{\bprob}{\begin{prob}}
\newcommand{\eprob}{\end{prob}}

\newcommand{\bvtm}{\begin{verbatim}}
\newcommand{\bfig}{\begin{figure}}
\newcommand{\efig}{\end{figure}}
\newcommand{\bcen}{\begin{center}}
\newcommand{\ecen}{\end{center}}

\long\def\comment#1{}

\def \n2{{N_0 \over 2}}

\def \h5{\hspace{0.5in}}

\newcommand{\dff}{\stackrel{\triangle}{=}}